%% file: neurips_2025.tex
\documentclass{article}

\usepackage{colortbl}
\usepackage[dvipsnames]{xcolor}
\usepackage{caption}
\usepackage{placeins}

\usepackage[final]{neurips_2025}
\usepackage{marvosym}
\usepackage{pifont}
\newcommand{\xmark}{\ding{55}}
\newcommand{\checkmark}{\ding{51}}

\usepackage[utf8]{inputenc} 
\usepackage[utf8]{inputenc} 
\usepackage{newunicodechar}
\newunicodechar{，}{,}
\usepackage[T1]{fontenc}    
\usepackage{pifont}
\usepackage{url}            
\usepackage{booktabs}       
\usepackage{amsfonts}       
\usepackage{nicefrac}       
\usepackage{microtype}      
\usepackage{xcolor}         
\usepackage{listings}
\usepackage{graphicx}
\usepackage{array}
\usepackage{amsmath}
\usepackage{amssymb}
\usepackage{adjustbox}
\usepackage{tcolorbox}
\usepackage{multirow}
\usepackage{makecell} 
\usepackage{tabularx}

\usepackage{booktabs}
\usepackage{multicol}
\usepackage{array}
\usepackage{xcolor}
\usepackage{colortbl}

\definecolor{RoboTwincolor1}{HTML}{65a487} 
\definecolor{RoboTwincolor2}{HTML}{68349a} 

\usepackage[breaklinks=true,
            colorlinks,
            linkcolor = RoboTwincolor1,
            urlcolor  = RoboTwincolor1, 
            citecolor = teal,
            bookmarks = false]{hyperref}

\title{
      \textbf{\textcolor{RoboTwincolor1}{Robo}\textcolor{RoboTwincolor2}{Twin} 2.0:} 
  A Scalable Data Generator and Benchmark with Strong Domain Randomization for Robust Bimanual Robotic Manipulation
}
\author{%
     \vspace{-4em}
    \\
    \\
    \textbf{Tianxing Chen}$^{2,16}\textsuperscript{*} ^\dag$, \textbf{Zanxin Chen}$^{3,5}\textsuperscript{*}$, \textbf{Baijun Chen}$^{15}\textsuperscript{*}$, \textbf{Zijian Cai}$^{3,5}\textsuperscript{*}$, \textbf{Yibin Liu}$^{13}\textsuperscript{*}$,\\ \textbf{Zixuan Li}$^{5}\textsuperscript{*}$, \textbf{Qiwei Liang}$^{5}$, \textbf{Xianliang Lin}$^{5}$, \textbf{Yiheng Ge}$^{1}$, \textbf{Zhenyu Gu}$^{7,8}$, \textbf{Weiliang Deng}$^{3,11}$, \\\textbf{Yubin Guo}$^{7,9}$, \textbf{Tian Nian}$^{3,5}$, \textbf{Xuanbing Xie}$^{12}$, \textbf{Qiangyu Chen}$^{5}$, \textbf{Kailun Su}$^{5}$, \textbf{Tianling Xu}$^{10}$, \\ \textbf{Guodong Liu}$^{6,7}$, \textbf{Mengkang Hu}$^{2}$, \textbf{Huan-ang Gao}$^{6,16}$, \textbf{Kaixuan Wang}$^{2，16}$, \\ \textbf{Zhixuan Liang}$^{2,3 \dag}$, \textbf{Yusen Qin}$^{4,6}$,  \textbf{Xiaokang Yang}$^{1}$, \textbf{Ping Luo}$^{2, 14 {\text{\Letter}}}$, \textbf{Yao Mu}$^{1,3 {\text{\Letter}} \dag}$ \\ \\
    $^1$ MoE key Lab of Artificial Intelligence, AI Institute, SJTU$^{\ddag}$, $^2$ HKU MMLab$^{\ddag}$, \\$^3$ Shanghai AI Lab, $^4$D-Robotics, $^5$SZU, $^6$THU, $^7$TeleAI, $^{8}$FDU, $^{9}$USTC, $^{10}$SUSTech,\\$^{11}$SYSU, $^{12}$CSU, $^{13}$NEU,$^{14}$HKU-SH ICRC, $^{15}$NJU, $^{16}$Lumina EAI\\
    $\textsuperscript{*}$ Equal contribution \quad $^{\text{\Letter}}$ Corresponding authors \quad $^{\dag}$ Co-project leads \\$^{\ddag}$ Equally leading organizations\\
    \vspace{-3pt}\\
    Webpage:\textbf{\href{https://robotwin-platform.github.io/}{https://\textcolor{RoboTwincolor2}{robotwin-platform.github.io}}} \quad Doc: \href{https://robotwin-platform.github.io/doc/}{https://robotwin-platform.github.io/doc/}
}

\begin{document}

\maketitle

\vspace{-3em}

\begin{figure*}[ht] 
    \centering
    \includegraphics[width=0.9\linewidth]{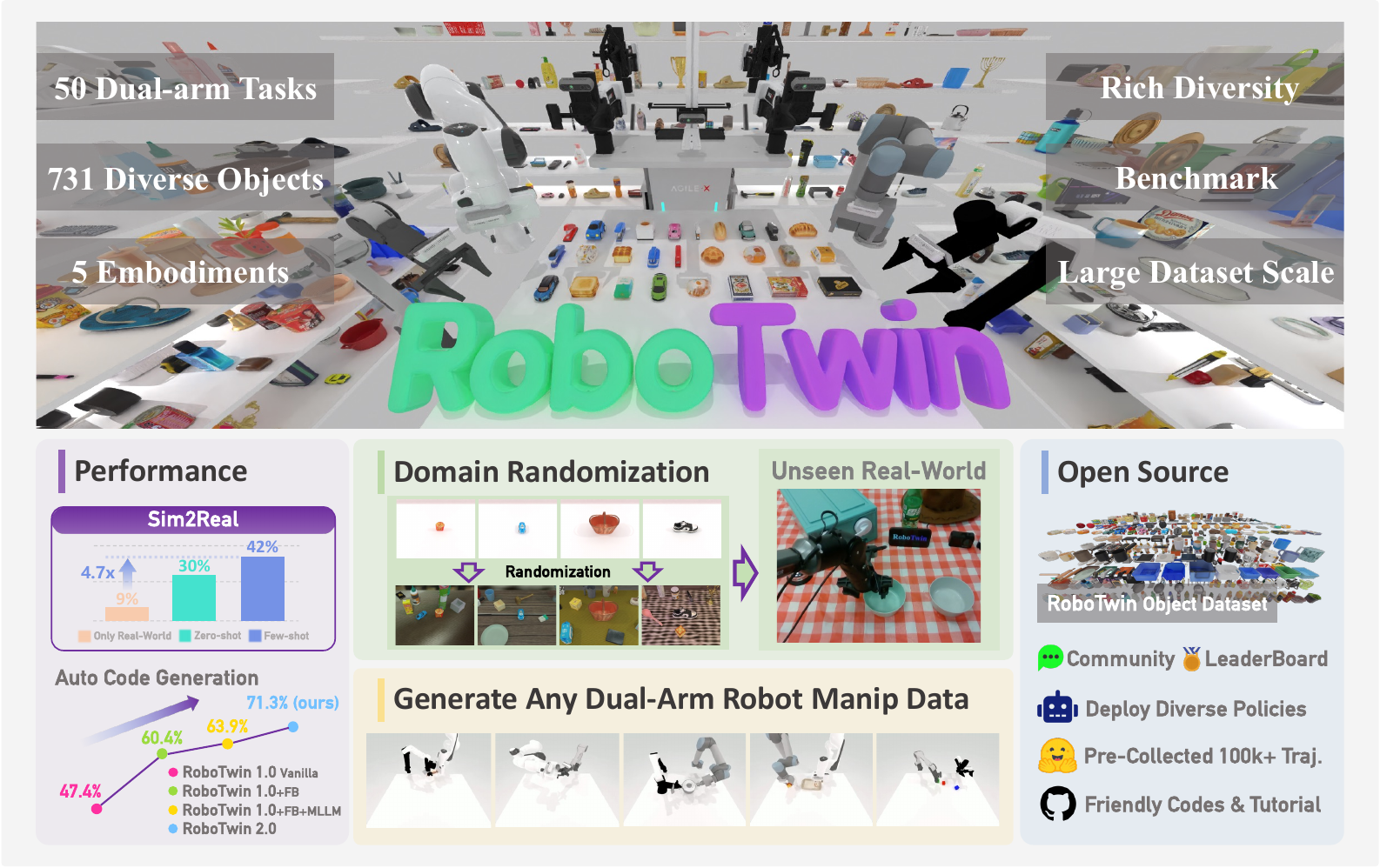}
    \caption{\textbf{Overview of \textcolor{RoboTwincolor1}{Robo}\textcolor{RoboTwincolor2}{Twin} 2.0.} RoboTwin 2.0 is a scalable framework for bimanual manipulation, integrating an expert data generation pipeline with a 50-task benchmark built on the RoboTwin Object Dataset (731 objects, 147 categories). A multimodal language agent automates task program synthesis, while flexible dual-arm configurations enable large-scale, diverse data collection. Policies trained on RoboTwin 2.0 exhibit improved robustness and generalization to unseen environments.}
    \label{fig:cover} 
\end{figure*}

\input{sections/abstract}

\input{sections/introduction}

\input{sections/method}

\input{sections/dataset}

\input{sections/experiment}

\input{sections/related_work}

\input{sections/conclusion}



\clearpage
{
\small
\bibliographystyle{plain}
\bibliography{ref}

}

\input{sections/appendix}

\end{document}

%% file: sections/abstract.tex
\begin{abstract}
Simulation-based data synthesis has emerged as a powerful paradigm for enhancing real-world robotic manipulation. However, existing synthetic datasets remain insufficient for robust bimanual manipulation due to two challenges: (1) the lack of an efficient, scalable data generation method for novel tasks, and (2) oversimplified simulation environments that fail to capture real-world complexity. We present \textbf{\textcolor{RoboTwincolor1}{Robo}\textcolor{RoboTwincolor2}{Twin} 2.0}, a scalable simulation framework that enables automated, large-scale generation of diverse and realistic data, along with unified evaluation protocols for dual-arm manipulation. We first construct RoboTwin-OD, a large-scale object library comprising 731 instances across 147 categories, each annotated with semantic and manipulation-relevant labels. Building on this foundation, we develop an expert data synthesis pipeline that combines multimodal large language models (MLLMs) with simulation-in-the-loop refinement to generate task-level execution code automatically. To improve sim-to-real transfer, RoboTwin 2.0 incorporates structured domain randomization along five axes: clutter, lighting, background, tabletop height and language instructions, thereby enhancing data diversity and policy robustness. We instantiate this framework across 50 dual-arm tasks spanning five robot embodiments. Empirical evaluation shows a 10.9\% gain in code generation success rate. Building on this, we evaluate downstream policy learning. With a mix of large-scale synthetic data and only 10 real demonstrations, a vision–language–action (VLA) model achieves a 367\% relative improvement over the 10-demo baseline. Even without real data, zero-shot models trained solely on synthetic data obtain a 228\% relative gain, highlighting the effectiveness of our dataset in strengthening sim-to-real transfer and robustness to environmental variations. We release the data generator, benchmark, pre-collected dataset, and code to support scalable research in robust bimanual manipulation.
\end{abstract}

%% file: sections/introduction.tex
\section{Introduction}

Bimanual robotic manipulation is critical for enabling robots to perform complex real-world tasks such as collaborative assembly, tool use, and object handovers. Developing generalizable bimanual policies—particularly vision–language–action (VLA) foundation models—requires datasets that are simultaneously high-quality, diverse, and large-scale. In the absence of sufficient variability in object geometry, scene clutter, lighting conditions, instruction language, and robot embodiments, learned policies often overfit to narrow distributions and fail to generalize to novel environments or hardware platforms. Yet collecting real-world demonstrations at scale remains prohibitively expensive, time-consuming, and logistically challenging, especially when aiming to cover a broad range of tasks, objects, and embodiments.

Simulation-based data generation provides a scalable alternative for collecting large-scale multimodal datasets and has shown promise in enabling sim-to-real transfer~\cite{mu2025robotwin,graspvla}. However, existing pipelines fall short in three critical aspects. First, they lack automated quality control: without an expert-level validation loop, many generated trajectories include execution failures or suboptimal grasps, which degrade policy learning. Second, their domain randomization is often superficial, yielding overly clean and homogeneous scenes that omit essential real-world factors such as clutter, lighting variation, and ambiguous language instructions—elements crucial for robust sim-to-real transfer. Third, they overlook cross-embodiment variation: different bimanual platforms can differ substantially in their kinematic capabilities and grasp strategies. For example, a low-degree-of-freedom (DoF) platform like the Piper often relies on lateral grasps due to its limited dexterity, whereas a high-DoF arm such as the Franka is capable of top-down precision grasps. Yet, current synthetic datasets rarely encode such embodiment-specific affordances or task constraints, limiting their generality.

To address these challenges, we introduce \textbf{\textcolor{RoboTwincolor1}{Robo}\textcolor{RoboTwincolor2}{Twin} 2.0}, a scalable simulation-based data generation framework designed to produce high-quality, diverse, realistic, and interaction-rich datasets for bimanual manipulation. RoboTwin 2.0 integrates three key components: (1) an automated expert data generation pipeline that leverages multimodal large language models (MLLMs) and simulation-in-the-loop feedback to iteratively validate and refine task execution code; (2) comprehensive domain randomization over language instructions, object clutter, background textures, lighting conditions, and tabletop configurations, aimed at closing the sim-to-real gap and enhancing policy generalization; and (3) embodiment-aware adaptation, in which object affordances are annotated and robot-specific action candidates are generated to account for heterogeneous dual-arm kinematics.

Building on these components, we introduce three new resources to support scalable research in bimanual manipulation: (1) the RoboTwin-OD asset library, comprising 731 annotated object instances across 147 categories; (2) an automated data generation pipeline with comprehensive domain randomization and a collection of over 100,000 expert trajectories spanning 50 tasks across five dual-arm robot platforms; and (3) a benchmark for evaluating policy generalization to cluttered environments and open-ended language goals. Together, these resources enable the community to train and evaluate robust bimanual manipulation policies under conditions that closely reflect real-world complexity and diversity.

In summary, our main contributions are as follows: (1) We develop an automated expert data generation framework that integrates multimodal large language models with simulation-in-the-loop feedback to ensure high-quality, expert-level trajectories;
(2) We propose a systematic domain randomization strategy that enhances policy robustness by increasing data diversity and sim-to-real generalization;
(3) We introduce an embodiment-aware adaptation mechanism that generates robot-specific manipulation candidates based on object affordances;
(4) We release the RoboTwin-OD asset library, a large-scale pre-collected multi-embodiment domain-randomized trajectory dataset, a scalable bimanual data generator, and a standardized evaluation benchmark to support scalable training and evaluation of generalizable policies across different robot embodiments, scene configurations, and language instructions.

%% file: sections/method.tex
\section{Method}

\begin{figure}[h] 
    \vspace{-10pt}
    \centering    \includegraphics[width=1.0\linewidth]{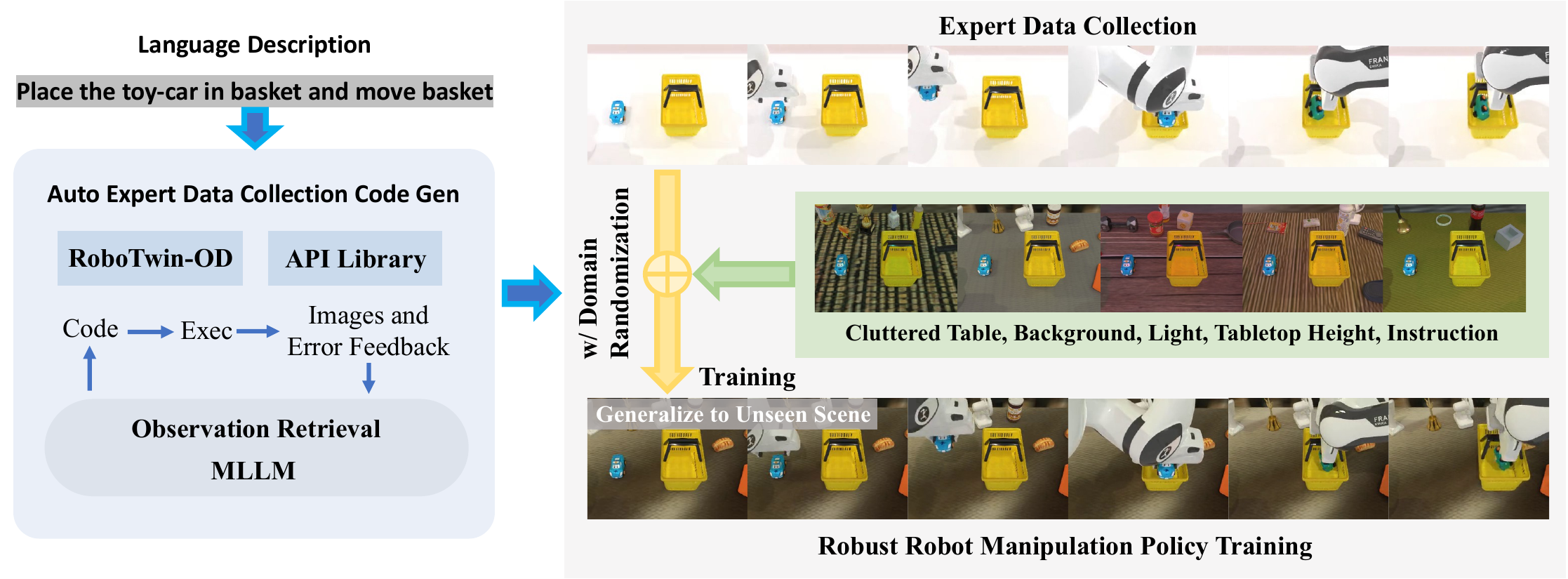}
    \vspace{-5pt}
    \caption{\textbf{RoboTwin 2.0 Pipeline.}
Built on RoboTwin-OD and a skill API, the framework uses MLLM-based code generation with simulation feedback to produce expert task programs and domain-randomized trajectories for policy training and evaluation.}
    \label{fig:RoboTwin_Pipeline} 
\end{figure}

\vspace{-0.5em}

We illustrate the overall RoboTwin 2.0 pipeline in Fig.~\ref{fig:RoboTwin_Pipeline}. The framework begins with a task code generation module that leverages multimodal large language models (MLLMs) and simulation-in-the-loop feedback to automatically synthesize executable task plans from natural language instructions. This module is grounded on a large-scale object asset library (RoboTwin-OD) and a predefined skill library, enabling scalable task instantiation across a broad range of object categories and manipulation scenarios. To ensure high-quality expert demonstrations, we integrate this automated generation pipeline with RoboTwin 2.0’s comprehensive domain randomization scheme, which diversifies observations along language, visual, and spatial axes. This pipeline supports the synthesis of diverse and realistic training data, facilitating the development of manipulation policies that are robust to real-world environmental variability.

\subsection{Expert Code Generation via MLLMs and Simulation-in-the-Loop Feedback}
\label{section3.2expert-data-gen}



Recent advances in language models show strong ability to generate intermediate task representations—such as textual plans~\cite{hu2024hiagent}, API calls, or executable code~\cite{mu2024robocodex,roboscript,hu2025text2world}—for complex robotic tasks. Multimodal large language models (MLLMs) further extend this capability by incorporating visual and proprioceptive inputs, enabling more grounded reasoning over real-world sensory data. However, prior systems often depend on strong manual priors or lack closed-loop feedback during program synthesis, which limits their robustness in diverse or dynamic environments.

\begin{figure}[h] 
    \vspace{-8pt}
    \centering 
    \includegraphics[width=0.9\linewidth]{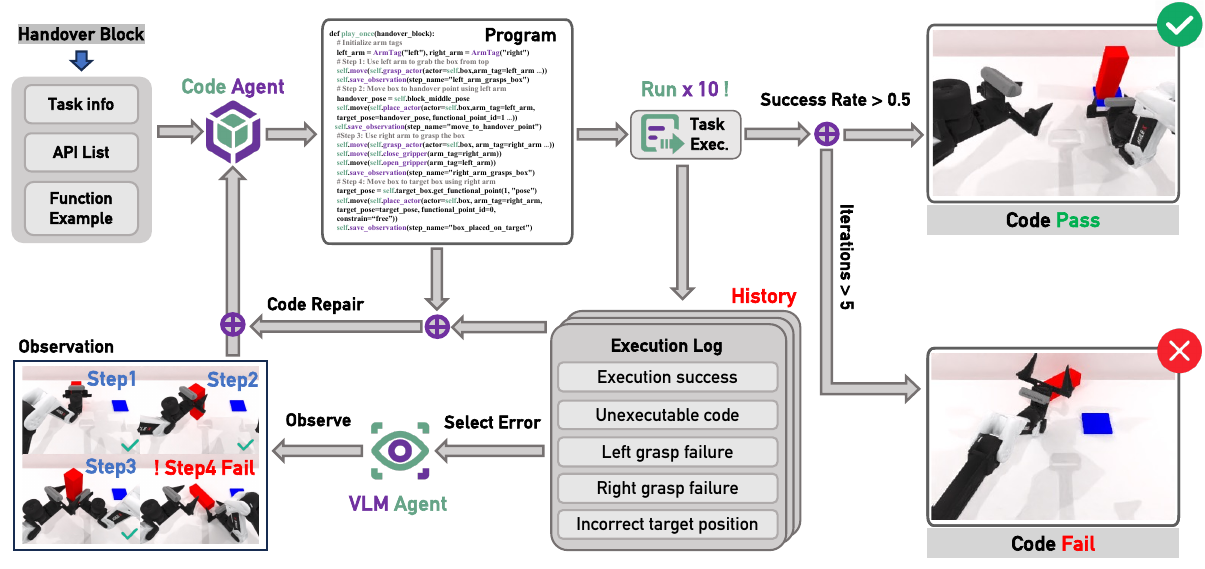}
    \caption{\textbf{Expert Code Generation Pipeline.} }
     \vspace{-8pt}
    \label{fig:gpt_gen} 
\end{figure}

To address these limitations, we propose an automated expert data generation pipeline that integrates programmatic code synthesis with multimodal execution feedback (Fig.\ref{fig:gpt_gen}). The system adopts a closed-loop architecture with two agents: a code-generation agent and a vision–language model (VLM) observer. The code agent synthesizes task programs from instructions, while the observer monitors execution in simulation, detects failures, and suggests corrections. This iterative feedback loop enables the code agent to refine programs automatically, producing robust, self-improving expert data with minimal human supervision. Unlike prior MLLM-based pipelines such as GenSim2\cite{huagensim2} and RoboGen~\cite{wang2023robogen}, our system supports zero-shot generation of complex dual-arm behaviors beyond primitive pick-and-place actions.

\label{section3.3domain-randomization}

\textbf{Input Specification.} Each task is defined by a task name (e.g., \textit{Handover Block}) and a natural language description of the objective. The code-generation agent is conditioned on three key inputs: a general API list, a set of example function calls, and a hierarchical constraint specification. These components collectively guide the synthesis of Python code to execute the task. Additionally, each task may include task-specific function call examples to further ground code generation in context.

\textbf{Initial Code Generation.} The code-generation agent synthesizes an initial Python program conditioned on the provided task inputs. It models the program synthesis process as a structured prediction problem over the space of available API calls, leveraging natural language understanding and few-shot prompting from task-specific examples. The generated code specifies a stepwise sequence of robot actions designed to accomplish the target manipulation objective.

\textbf{Simulated Execution and Logging.} The generated program is executed ten times per iteration within a simulated robotic environment. Multiple trials are used to account for stochastic variations in simulation dynamics, robot controllers, and sensor noise. After each execution batch, the system generates a structured execution log that records the success or failure of each trial and annotates failure cases with their corresponding causes—such as unexecutable code, left/right grasp failure, or incorrect object placement.

\textbf{Multimodal Observation and Error Localization.} In parallel with execution, a vision-language model (VLM) agent observes the robot’s behavior across all ten trials. The VLM performs frame-by-frame inspection to evaluate the success of each program step and localize the point of failure when errors occur. Beyond temporal localization, the VLM also diagnoses failure modes by inferring whether the underlying cause stems from flawed logic, incorrect API usage, or other systemic issues. This diagnostic capability enables the system to address root causes rather than merely responding to superficial execution errors. The detail of VLM observation is shown in \ref{app:vlm_observe}.

\textbf{Code Repair and Iterative Refinement.} The code-generation agent receives two complementary feedback signals: (i) a quantitative execution log and (ii) a qualitative, localized diagnostic from the VLM. It integrates these inputs to revise the program by modifying or replacing instructions identified as failure-prone. The updated program is then re-evaluated in the next iteration, and the process continues until either the program achieves the setting success rate across ten simulated runs in one iteration or fails to do so after five consecutive refinements. This loop yields expert-level task code with minimal human supervision while avoiding indefinite refinement.

The outcome of this pipeline is a collection of robust, automatically synthesized programs that generate high-quality expert trajectories for downstream training and evaluation. By integrating multimodal reasoning with execution-level feedback, the system produces code that is not only syntactically correct but also semantically aligned with task objectives. This closed-loop generation framework substantially reduces human supervision while enabling scalable and self-improving expert data creation for complex robotic manipulation tasks.

\subsection{Domain Randomization for Robust Robotic Manipulation}

\begin{figure}[h] 
    \centering
    \includegraphics[width=1.0\linewidth]{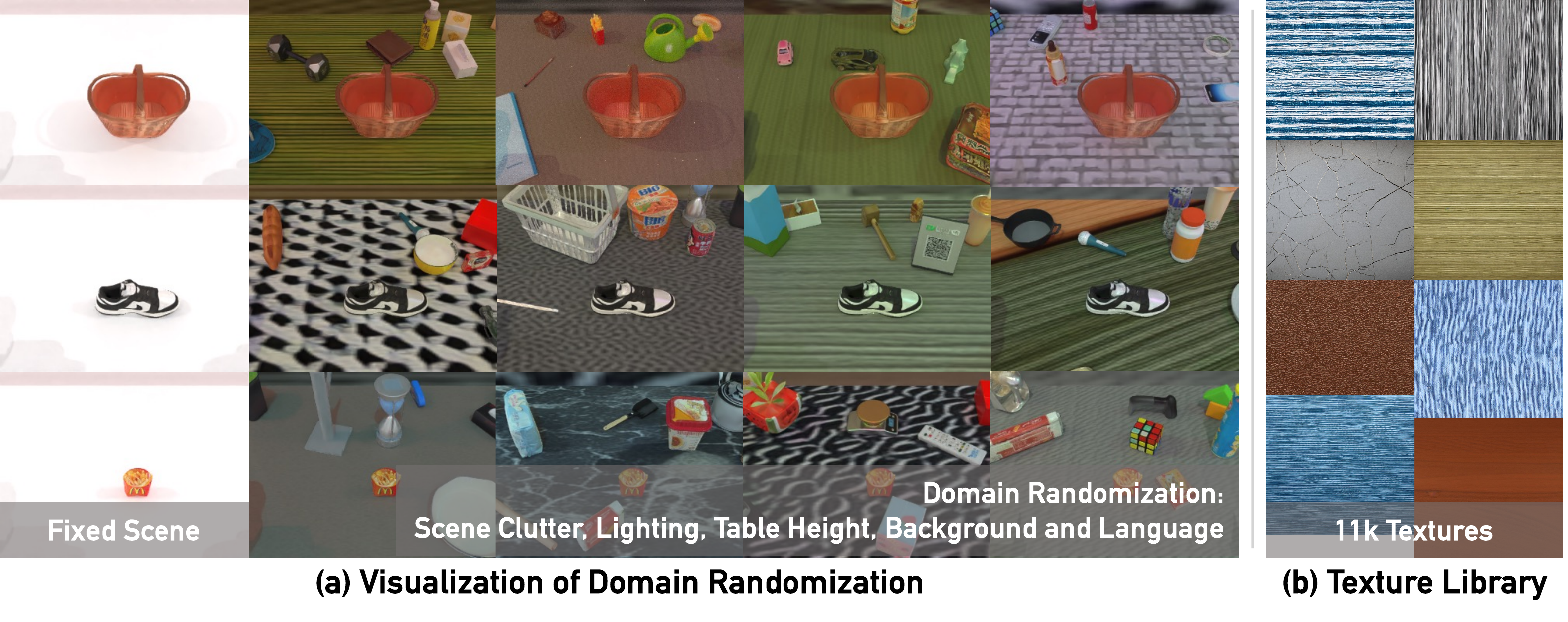}
    \vspace{-15pt}
    \caption{\textbf{Visualization of domain randomization and our texture library.} }
    \vspace{-5pt}
    \label{fig:vis_randomization} 
\end{figure}

To enhance policy robustness to real-world variability, we apply domain randomization along five dimensions: (1) cluttered distractor objects, (2) background textures, (3) lighting conditions, (4) tabletop heights, and (5) diverse language instructions. This systematic augmentation broadens the training distribution and markedly improves generalization to unseen scenarios. The effects of these randomizations are visualized in Fig.~\ref{fig:vis_randomization}a.

\textbf{Scene Clutter.} To enhance robustness to environmental variation, we augment tabletop scenes with task-irrelevant distractors drawn from RoboTwin-OD (731 objects across 147 categories; see Section~\ref{robotwin-od}). Each object includes placement annotations, enabling a generic API for semantically valid insertion. We ensure physical plausibility via collision-aware placement and precomputed volumes. To avoid policy confusion, distractors visually or semantically similar to task-relevant objects are excluded during sampling. This yields diverse yet unambiguous cluttered scenes for training.


\textbf{Diverse Background Textures.} We randomize tabletop surfaces and backgrounds using a large curated texture library. To build it, we first collected 1,000 diverse surface descriptions via LLM prompting and web crawling, then used Stable Diffusion v2 to generate 20 samples per description (20,000 total). After human-in-the-loop filtering, we obtained 11,000 high-quality textures. This library is applied in simulation to enrich visual diversity and reduce overfitting to clean synthetic environments (see Fig.~\ref{fig:vis_randomization}b).

\textbf{Lighting Variation.} Real-world environments exhibit diverse illumination conditions, with variations in color temperature, source type, number, and placement. These factors alter object appearance and reflections, challenging vision-based manipulation. To enhance robustness, we randomize light color, type, intensity, and position within physically plausible bounds. As shown in Fig.~\ref{fig:vis_randomization}a (second row), changes in color temperature can drastically shift object appearance (e.g., a shoe under warm vs. cool light). Training under such randomized conditions improves policy robustness to real-world illumination shifts.

\textbf{Tabletop Heights.} In practice, table heights vary across workspaces, affecting robot perception, kinematics, and interaction. To improve generalization, we uniformly randomize table height within a plausible range during simulation, introducing variability in viewpoints and spatial relations between robot and objects.


\textbf{Trajectory-Level Diverse Language Instructions.} To improve robustness to natural language variation, we use a multimodal LLM to generate diverse task templates and multiple object descriptions capturing geometry, appearance, and part-level attributes. Each task and object thus has several alternative phrasings, which can be flexibly combined. For every trajectory, we sample from these pools to compose instructions. For example, in \textit{Move Can Pot}, the template “Use {a} to place {A} to the left of {B}” may yield diverse instructions such as “Use left arm to place sauce can to the left of gray kitchenpot” or “Use left arm to place white plastic lid sauce can to the left of kitchenpot for boiling and cooking.” This combinatorial augmentation produces a large set of linguistically varied instructions and significantly improves generalization to unseen language and scene configurations (see Appendix~\ref{fig:diverse-language-demo}, \ref{prompt-description}).

\subsection{Embodiment-Aware Grasp Adaptation}
\label{grasping-adaptation}

\begin{figure*}[h]
    \vspace{-0.5em}
  \centering
  \begin{minipage}{0.71\textwidth}
    \centering
    \includegraphics[width=0.99\linewidth]{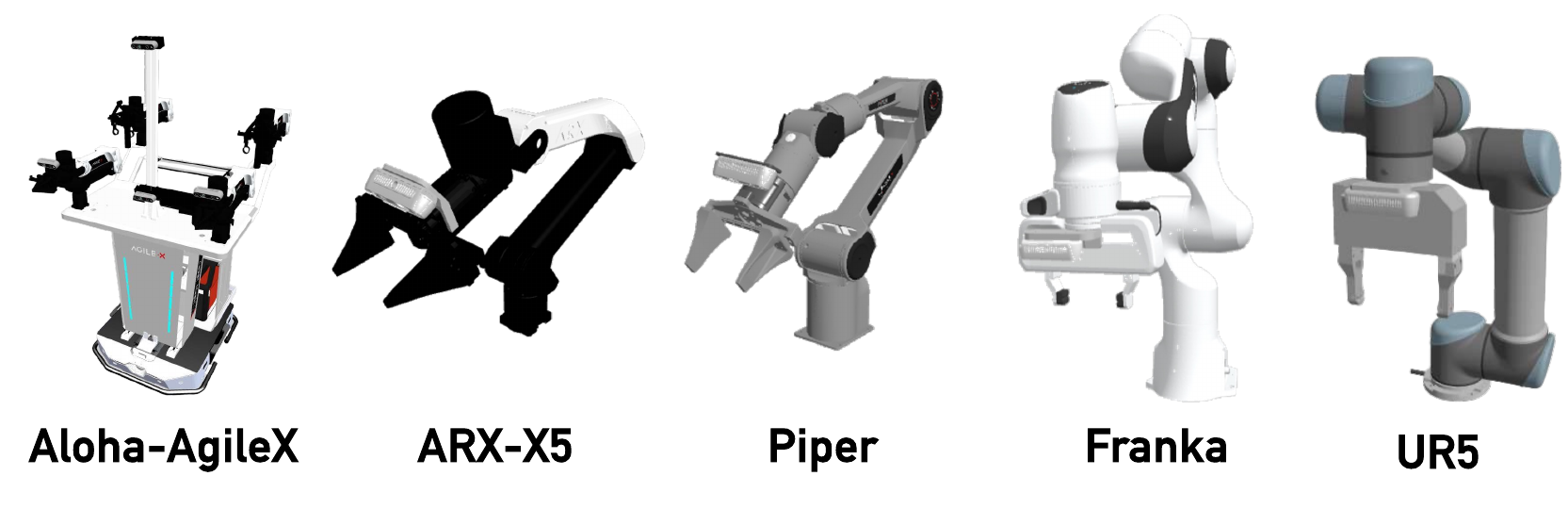}
    \vspace{-0.7em}
    \captionof{figure}{\textbf{Five RoboTwin 2.0 Embodiments.}}
    \label{fig:available-embodiments} 
  \end{minipage}
  \hfill
  \begin{minipage}{0.27\textwidth}
    \centering
    \vspace{5pt}
    \includegraphics[width=0.95\linewidth]{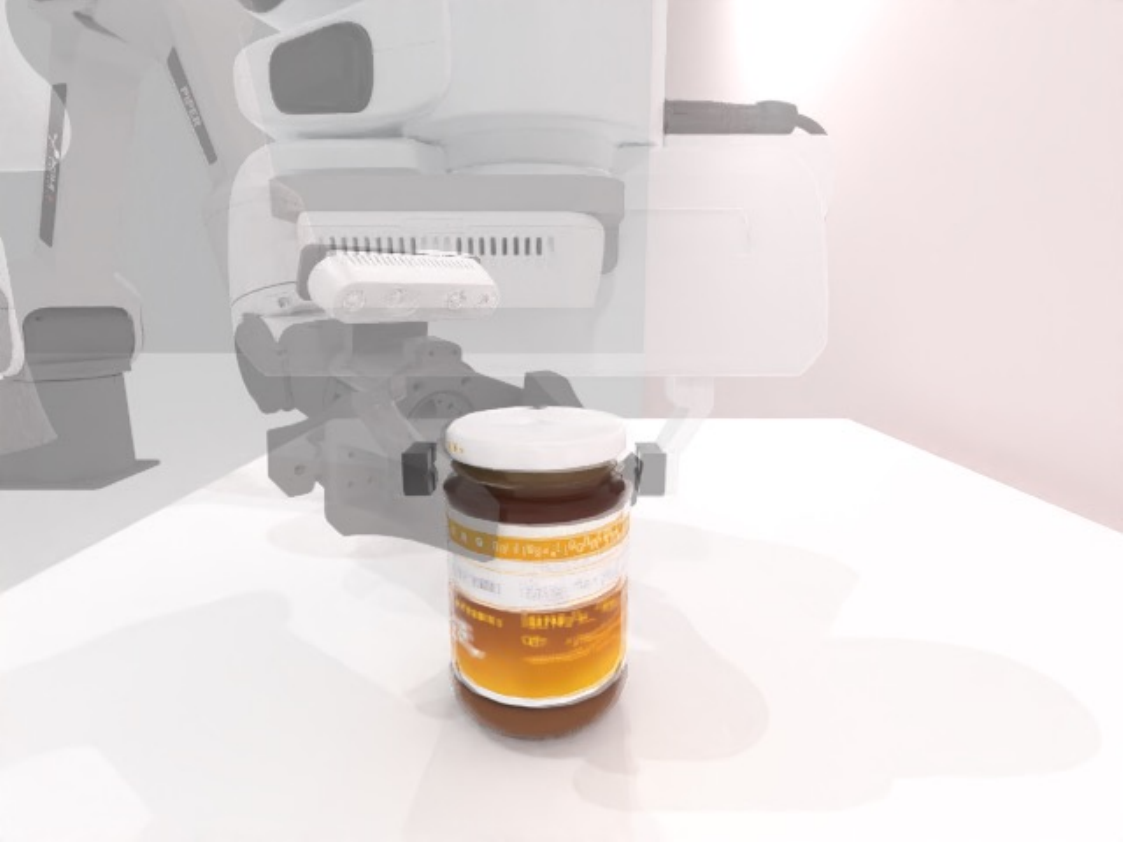}
    \vspace{1pt}
    \captionof{figure}{\textbf{Different Grasping Behavior.}}
    \label{fig:diverse-grasp} 
  \end{minipage}
\end{figure*}

Due to differences in DoF and kinematic structures, robotic arms exhibit varying reachable workspaces and preferred manipulation strategies for the same task. For example, when grasping a can, the Franka arm typically favors a top-down approach, while the lower-DoF Piper arm is better suited to side grasps. As a result, a task successfully completed by Franka using a top-down grasp may require a side approach when executed with Piper, as shown in Fig.~\ref{fig:diverse-grasp}.

To address embodiment-specific variations, we annotate each object with a rich set of candidate manipulation poses that cover multiple grasp axes and approach directions. This design captures both manipulation diversity and robot-specific preferences. To further expand the feasible space, we apply angular perturbations biased toward directions with higher arm reachability. Concretely, for each object we generate candidate grasps by combining preferred operation directions, randomized pose perturbations, and parallelized motion planning attempts.

%% file: sections/dataset.tex
\section{RoboTwin 2.0 Data Generator, Benchmark and Large Scale Dataset}

\subsection{RoboTwin-OD: RoboTwin Object Dataset}
\label{robotwin-od}

\begin{figure}[h] 
    \centering
\includegraphics[width=0.9\linewidth]{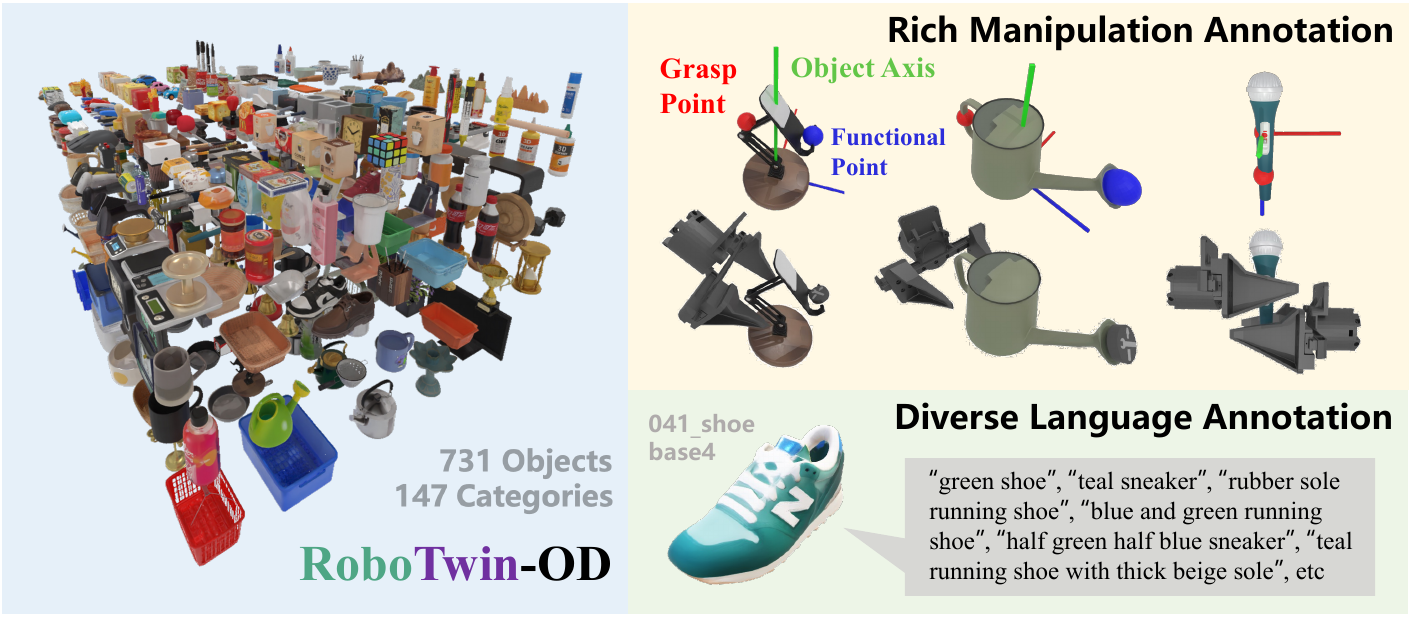}
    \vspace{-5pt}
    \caption{\textbf{RoboTwin-OD.} A large-scale object dataset for robotic manipulation with 147 categories and 731 objects, annotated with rich interaction labels and diverse language descriptions.}
     \vspace{-5pt}
    \label{fig:object-library} 
\end{figure}

To enhance both manipulation capability and visual understanding, we construct a large-scale object dataset with rich semantic annotations, called RoboTwin-OD, covering 147 categories and 731 diverse objects. Specifically, this includes 534 instances across 111 categories that we generated in-house using RGB-to-3D reconstruction via the Rodin platform\footnote{\url{https://hyper3d.ai/}}, followed by convex decomposition and mesh merging to ensure physically accurate collision models. In addition, RoboTwin-OD incorporates 153 objects from 27 categories in Objaverse~\cite{deitke2023objaverse}, and 44 articulated object instances from 9 categories in SAPIEN PartNet-Mobility~\cite{xiang2020sapien}. Objects from all sources, including Objaverse, are used to construct cluttered scenes, with Objaverse specifically enhancing the visual and semantic diversity of distractor objects. We also develop a comprehensive texture library for surfaces and backgrounds using generative AI and human-in-the-loop filtering to ensure visual realism and diversity.

For robust manipulation, policies must generalize across diverse objects, which requires datasets with broad category coverage and varied intra-class instances. To facilitate language grounding, we developed an automated object description generator with human verification, producing 15 annotations per object that vary in shape, texture, functionality, part structure, and granularity.

To further support object-centric interaction, we annotate each object with key point–axis information, including placement points, functional points, grasp points, and grasp axes, explicitly encoding affordances. Together with our manipulation API library, these annotations enable generalizable grasp execution in simulation. All object information is available at \href{http://robotwin-platform.github.io/doc/objects}{http://robotwin-platform.github.io/doc/objects/}
.



\subsection{50 Tasks for Data Generation and Benchmarking}
\label{all_tasks}

\begin{figure}[h] 
    \centering    
    \includegraphics[width=1.0\linewidth]{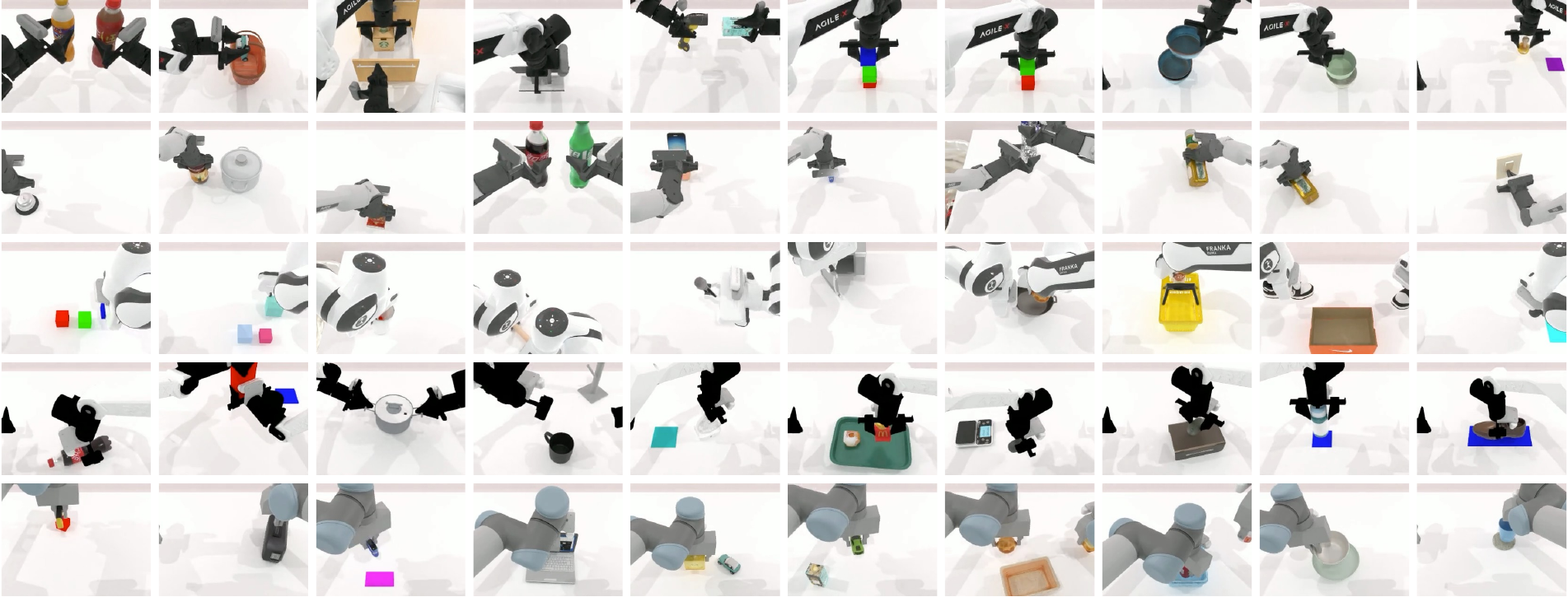}
    \vspace{-7pt}
    \caption{\textbf{50 RoboTwin 2.0 Bimanual Manipulation Tasks}.}
    \label{fig:RoboTwin_tasks} 
    \vspace{-5pt}
\end{figure}

Building on our automated task generation framework, embodiment-adaptive behavior synthesis, and the large-scale RoboTwin-OD asset library, we construct a suite of 50+ dual-arm collaborative manipulation tasks. We further support data collection and evaluation on five distinct robot platforms, enabling comprehensive cross-embodiment benchmarking. Keyframes from representative tasks are shown in Fig.~\ref{fig:RoboTwin_tasks}, and the complete task descriptions are available at \href{http://robotwin-platform.github.io/doc/tasks}{http://robotwin-platform.github.io/doc/tasks/}. We also pre-collected over 100,000 dual-arm manipulation trajectories across 50 tasks in RoboTwin 2.0, which are available at \href{https://huggingface.co/datasets/TianxingChen/RoboTwin2.0/tree/main/dataset}{https://huggingface.co/datasets/TianxingChen/RoboTwin2.0/tree/main/dataset}.

%% file: sections/experiment.tex
\section{Experiment}
\label{experiment}

We design experiments to evaluate the effectiveness of RoboTwin 2.0 in three key aspects:
(1) automating the generation of high-quality expert code for manipulation tasks;
(2) improving policy robustness to environmental variation via diversified training data; and
(3) demonstrating the utility and diversity of RoboTwin 2.0 as a standardized benchmark for evaluating policy generalization across tasks, scenes, and embodiments.

\subsection{Evaluation of Automated Expert Code Generation}

We evaluate our closed-loop expert data generation system on a suite of 10 robotic manipulation tasks, each specified with a natural language instruction. For each configuration, the code-generation agent produces multiple candidate programs, which are executed in simulation to account for stochasticity in dynamics, control, and perception. Task-level success is defined as the average success rate across all executions, as described in Section~\ref{section3.2expert-data-gen}.

\begin{figure*}[h]
  \centering
  \begin{minipage}{0.58\textwidth} 
    \centering
    \scriptsize 
\captionof{table}{\textbf{Overall performance comparison across RoboTwin variants.} Evaluated on the subset of tasks supported by both RoboTwin 1.0 and RoboTwin 2.0. Per-task success rate comparison is provided in Appendix~\ref{app:task_performance}.}
    \vspace{2pt}
    \setlength{\tabcolsep}{4.5pt} 
    \renewcommand{\arraystretch}{1.1} 
    \begin{tabular}{lcccc}
      \toprule
      \textbf{Method} & \textbf{ASR} & \textbf{Top5-ASR} & \textbf{CR-Iter} & \textbf{Token} \\
      \midrule
      R1.0 Vanilla     & 47.4\% & 57.6\% & 1.00 & 1236.6 \\
      R1.0 + FB             & 60.4\% & 71.4\% & 2.46 & 1190.4 \\
      R1.0 + MM FB     & 63.9\% & 74.2\% & 2.42 & 1465.0 \\
      \midrule
      R2.0 Vanilla     & 62.1\% & 68.0\% & 1.00 & \textbf{569.4} \\
      R2.0 + FB            & 66.7\% & 73.6\% & 1.89 & 581.6 \\
      R2.0 + MM FB     & \textbf{71.3\%} & \textbf{78.6\%} & \textbf{1.76} & 839.7 \\
      \bottomrule
    \end{tabular}
    \label{tab:robotwin_compare_results}
  \end{minipage}
  \hfill
  \begin{minipage}{0.4\textwidth} 
    \centering
    \includegraphics[width=1.0\linewidth]{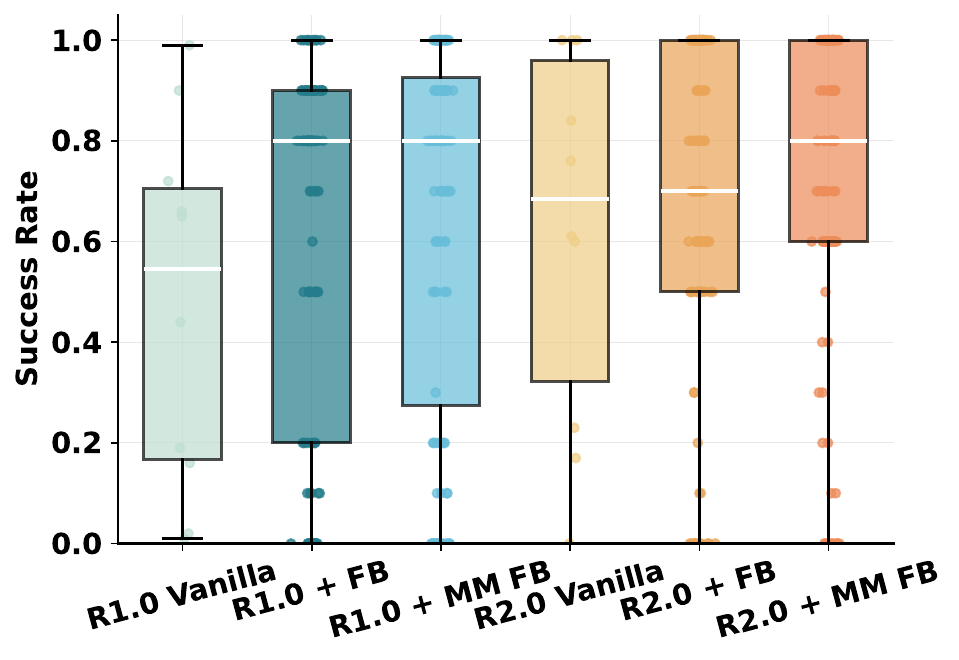}
    \caption{\textbf{RoboTwin Success Rate Distribution.}}
    \label{fig:robotwin-success-rate-distribution} 
  \end{minipage}
\end{figure*}

We evaluate performance with four metrics: \textbf{ASR} (Average Success Rate), \textbf{Top5-ASR} (success over the top-5 candidates per task), \textbf{CR-Iter} (average refinement iterations before termination), and Token (average number of tokens in generated policy code). Results on RoboTwin 1.0 and 2.0 are reported in Table~\ref{tab:robotwin_compare_results} under three configurations: \textit{Vanilla} (one-shot code generation), \textit{FB} (feedback-based repair via execution logs), and \textit{MM FB} (multimodal feedback with vision–language diagnostics). Per-task success rates are provided in Appendix~\ref{app:task_performance}.

Across all settings, multimodal feedback yields consistent gains. In RoboTwin 1.0, ASR improves from $47.4\%$ (Vanilla) to $63.9\%$ (MM FB); in RoboTwin 2.0, it rises from $62.1\%$ to $71.3\%$. Improvements are also evident in Top5-ASR, suggesting that perceptual feedback disproportionately benefits the best candidate programs. RoboTwin 2.0 converges faster than 1.0 (e.g., $1.76$ vs.\ $2.42$ CR-Iter in MM FB), indicating stronger priors and more efficient refinement. Token cost is also substantially reduced, especially in Vanilla (569.4 vs.\ 1236.6), reflecting more concise initial code.

Figure~\ref{fig:robotwin-success-rate-distribution} further shows that feedback narrows the success-rate distribution and raises the median. RoboTwin 2.0 with multimodal feedback achieves compact distributions centered above 80\%, highlighting robustness and reliability.

Overall, three findings emerge: (1) vision–language feedback not only detects failures but also guides precise repairs; (2) architectural improvements in RoboTwin 2.0 accelerate convergence and reduce token usage; and (3) combining symbolic execution logs with perceptual diagnostics yields more reliable, semantically aligned expert data. Together, these results validate the effectiveness of our closed-loop, self-improving code generation architecture. Detailed setups, metric definitions, and additional analyses are provided in Appendix~\ref{code_gens}.

\subsection{Evaluating Efficiency with and without Adaptive Grasping}

\begin{figure*}[h]
    \centering
    \footnotesize
    \captionof{table}{\textbf{Overall Performance Comparison between RoboTwin 1.0 and RoboTwin 2.0.}}
    \vspace{2pt}
  \resizebox{0.8\textwidth}{!}{%
    \begin{tabular}{lccccc|c}
      \toprule
      \textbf{Method} & \textbf{Aloha-AgileX} & \textbf{Piper} & \textbf{Franka} & \textbf{UR5} & \textbf{ARX-X5} & \textit{\textbf{Average}} \\
      \midrule
      RoboTwin 1.0 & $65.1\%$ & $2.4\%$ & \cellcolor{blue!15}\textbf{67.3}\% & \cellcolor{blue!15}\textbf{57.6}\% & $68.6\%$ & $52.2\%$ \\
      RoboTwin 2.0 & \cellcolor{blue!15}\textbf{78.8}\% & \cellcolor{blue!15}\textbf{25.1}\% & $67.2\%$ & $57.1\%$ & \cellcolor{blue!15}\textbf{74.2}\% & \cellcolor{blue!15}\textbf{60.5}\% \\
      \midrule 
      \textit{Difference} & \textcolor{ForestGreen}{\textbf{+13.7\%}} & \textcolor{ForestGreen}{\textbf{+22.7\%}} & \textcolor{red}{\textbf{-0.1\%}} & \textcolor{red}{\textbf{-0.5\%}} & \textcolor{ForestGreen}{\textbf{+5.6\%}} & \textcolor{ForestGreen}{\textbf{+8.3\%}} \\
      \bottomrule
    \end{tabular}
    }
    \label{tab:grasp-success}
\end{figure*}
To evaluate the effectiveness of our embodiment-aware grasp augmentation strategy, we measure the task success rate of automated data collection across 50 RoboTwin 2.0 tasks on five different robot embodiments. As shown in Table~\ref{tab:grasp-success}, we compare our RoboTwin 2.0 pipeline against the RoboTwin 1.0 baseline, which lacks diverse grasping and candidate augmentation. Results show that our method improves success rates, particularly for robots with constrained planning spaces, achieving an average improvement of 8.3\% across all embodiments. Specifically, for high-DoF arms with large reachable workspaces, such as Franka and UR5 (7-DoF), success rates remain largely unchanged, indicating limited benefit when the robot already has sufficient kinematic flexibility. However, for lower-DoF platforms such as Aloha-AgileX, Piper, and ARX-X5 (6-DoF), our method leads to substantial gains of 13.5\%, 22.7\%, and 5.6\%, respectively. These results demonstrate that our approach provides additional feasible grasp options that effectively mitigate the planning limitations of low-DoF manipulators. Success rates for all tasks can be found in Appendix~\ref{all_planning_success_rate}.

\subsection{Assessing the Impact of RoboTwin 2.0 on Policy Robustness}

\label{good-robustness}

Our goal is to evaluate whether the domain-randomized data in RoboTwin 2.0 can endow models with robustness to environmental perturbations. To this end, we first pre-train RDT and Pi0 on 9,600 expert trajectories collected from 32 tasks (300 per task) under two settings: clean (non-randomized) and domain-randomized.

For comparison, we also evaluate the released pretrained weights of RDT and Pi0 without additional fine-tuning. To further study generalization, we select five unseen tasks and collect 50 clean demonstrations per task for single-task training and fine-tuning. Finally, all policies—including ACT, DP, RDT, and Pi0—are evaluated under domain-randomized conditions to measure robustness in previously unseen environments. Detailed configurations are provided in Appendix~\ref{domain_randomization_setting} and~\ref{details}.


\begin{table*}[h]
  \centering
\footnotesize
\setlength{\tabcolsep}{5pt}
\caption{\textbf{Evaluating the Impact of RoboTwin 2.0 on Policy Robustness.} }
    \begin{tabular}{*{1}{>{\centering\arraybackslash}m{2.8cm}} *{4}{>{\centering\arraybackslash}m{0.65cm}} *{4}{>{\centering\arraybackslash}m{1.0cm}}}
    \toprule
       \textbf{\makecell[c]{Simulation\\Tasks}} & \textbf{ACT} & \textbf{DP} & \textbf{RDT} & \textbf{Pi0} & \textbf{\makecell[c]{RDT\\+Clean}} & \textbf{\makecell[c]{Pi0\\+Clean}} & \textbf{\makecell[c]{RDT\\+Rand.}} & \textbf{\makecell[c]{Pi0\\+Rand.}}\\
    \midrule
    {Stack Bowls Two} & 0.0\% & 0.0\% & 30.0\% & 41.0\%& 8.0\% & 55.0\% & 49.0\% &  62.0\% \\
    \midrule
    {Pick Dual Bottles} & 0.0\% & 0.0\% & 13.0\% & 12.0\% & 12.0\% & 15.0\% & 17.0\%  & 7.0\% \\
    \midrule
    {Move Can Pot} & 4.0\% & 0.0\% & 12.0\% & 21.0\% & 13.0\% & 35.0\% & 18.0\% & 22.0\%\\
    \midrule
    {Place Object Basket} & 0.0\% & 0.0\% & 17.0\% & 2.0\%  & 9.0\% & 8.0\% & 6.0\% & 22.0\% \\
    \midrule
    {Place Shoe} & 0.0\% & 0.0\% & 7.0\% & 6.0\% & 9.0\% & 6.0\% & 30.0\% & 18.0\% \\
    \midrule
    {Open Laptop} & 0.0\% & 0.0\% & 32.0\% & 46.0\% & 21.0\% & 33.0\% & 35.0\% & 50.0\% \\
    \midrule
    {Press Stapler} & 6.0\% & 0.0\% & 24.0\% & 29.0\% & 21.0\% & 26.0\% & 27.0\% & 31.0\% \\
    \midrule
    {Turn Switch} & 2.0\% & 1.0\% & 15.0\% & 23.0\% & 24.0\% & 21.0\% & 16.0\% & 21.0\% \\
    \midrule
    \textbf{\textit{Average}} & 2.0\% & 0.0\% & 18.8\% & 22.5\% & 14.6\% & 24.9\% & 24.8\% & 29.1\% \\
    \bottomrule
    \end{tabular}
  \label{tab:add_randomization-results}%
\end{table*}

As shown in Table~\ref{tab:add_randomization-results}, we observe that models fine-tuned with clean data show negligible improvements in average success rate compared to their pretrained counterparts, indicating that data without domain randomization does not help the model handle environmental variations. This also suggests that the low success rate of pretrained VLAs in simulation is not due to a Real-to-Sim gap, since we provide clean simulation data yet observe no clear improvement. In contrast, models pretrained with RoboTwin 2.0 data exhibit significantly improved generalization. Specifically, RDT and Pi0 achieve relative improvements of 31.9\% and 29.3\%. Notably, this performance gain persists even though the downstream tasks were trained using only clean, non-randomized data. This demonstrates that domain-randomized pretraining with RoboTwin 2.0 effectively equips models with robustness to visual and spatial variations. As a result, models pretrained with RoboTwin 2.0 can adapt to new tasks without requiring additional data augmentation or complex scene variations.

\subsection{Evaluation on Sim-to-Real Performance}
\label{real-world}

\begin{figure*}[h] 
    \centering    
    \includegraphics[width=1.0\linewidth]{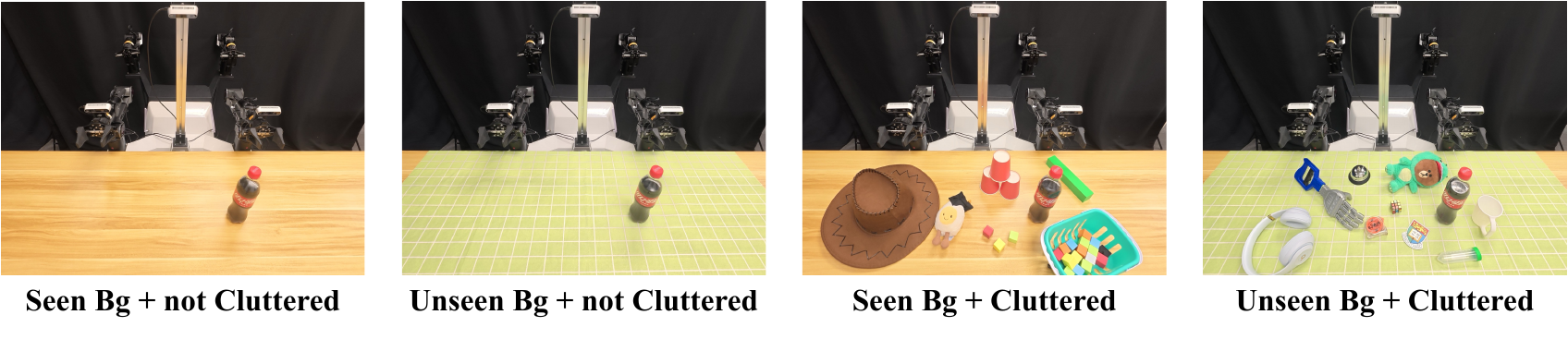}
    \vspace{-10pt}
    \caption{\textbf{Real-World Evaluation across Four Configurations.}}
    \label{fig:real-world-figure} 
    \vspace{-10pt}
\end{figure*}

To evaluate RoboTwin 2.0’s effectiveness in enhancing real-world policy robustness, we conduct experiments on four bimanual tasks: \textit{Stack Bowls}, \textit{Handover Block}, \textit{Pick Bottle}, and \textit{Click Bell}. All experiments use RDT as the policy backbone and are executed on the COBOT-Magic dual-arm platform. We compare three training settings: (1) 10 real-world demonstrations in clean tabletop environments; (2) the same demonstrations augmented with 1,000 domain-randomized synthetic trajectories generated under cluttered scenes with varied lighting and backgrounds; and (3) a synthetic-only setting trained solely on the 1,000 domain-randomized trajectories. To further improve robustness to camera jitter and calibration errors, we apply random 3D perturbations to simulated camera poses (position and orientation), with the displacement magnitude bounded by 1 cm.

Evaluation is conducted under four test configurations: (i) clean tabletop with seen backgrounds, (ii) clean tabletop with unseen backgrounds, (iii) cluttered tabletop with seen backgrounds, and (iv) cluttered tabletop with unseen backgrounds (Fig.\ref{fig:real-world-figure}). Since the synthetic-only setting excludes seen backgrounds during training, the corresponding entries in Table~\ref{tab:real-world-results} are omitted. This setup directly tests whether RoboTwin 2.0 enables robust policy generalization without additional real-world data from visually complex environments.

\begin{table*}[ht]
  \centering
  \footnotesize
  \setlength{\tabcolsep}{4pt}
  \caption{\textbf{Real-World Experiment Results.} We conduct controlled experiments on 4 dual-arm tasks: \textit{Stack Bowls}, \textit{Handover Block}, \textit{Pick Bottle}, and \textit{Click Bell}, each evaluated under 4 different settings.}
  \resizebox{0.9\textwidth}{!}{%

    \begin{tabular}{*{1}{>{\centering\arraybackslash}m{2cm}} *{2}{>{\centering\arraybackslash}m{1.6cm}} *{1}{>{\centering\arraybackslash}m{1.5cm}} *{2}{>{\centering\arraybackslash}m{2.2cm}}}
    \toprule
       \textbf{\makecell[c]{Real World\\Task}}  & \textbf{\makecell[c]{Background\\Type}} & \textbf{\makecell[c]{Cluttered\\or Not}}  & \textbf{10 Clean Real} & \textbf{\makecell[c]{10 Clean Real +\\1k RoboTwin 2.0}} & \textbf{\makecell[c]{1k RoboTwin 2.0\\(\textcolor{red}{Zero-shot})}} \\
    \midrule
    
    \multirow{4}{*}{\textit{Stack Bowls}} & 
        \multirow{2}{*}{Seen} &
            False & 22.0\% & \cellcolor{blue!15}\textbf{64.0}\%  & / \\ 
        & & True & 12.0\% & \cellcolor{blue!15}\textbf{58.0}\%  & / \\
        
        \cmidrule{2-6}
        
        & \multirow{2}{*}{Unseen} &
            False & 10.0\% & 50.0\% & \cellcolor{blue!15}\textbf{60.0}\% \\ 
        & & True & 12.0\% & \cellcolor{blue!15}\textbf{56.0}\% & 52.0\% \\
        
    \midrule
    
    \multirow{4}{*}{\textit{Handover Block}} & 
        \multirow{2}{*}{Seen} &
            False & 40.0\% & \cellcolor{blue!15}\textbf{48.0}\%  & / \\ 
        & & True & \cellcolor{blue!15}\textbf{16.0}\% & 12.0\%  & / \\
        
        \cmidrule{2-6}
        
        & \multirow{2}{*}{Unseen} &
            False & 36.0\% & \cellcolor{blue!15}\textbf{56.0}\% & \cellcolor{blue!15}\textbf{56.0}\%  \\ 
        & & True & 0.0\% & \cellcolor{blue!15}\textbf{36.0}\%  & 20.0\% \\
        
    \midrule
    
    \multirow{4}{*}{\textit{Pick Bottle}} & 
        \multirow{2}{*}{Seen} &
            False & 20.0\% & \cellcolor{blue!15}\textbf{36.0}\%  & / \\ 
        & & True & 8.0\% & \cellcolor{blue!15}\textbf{40.0}\%  & / \\
        
        \cmidrule{2-6}
        
        & \multirow{2}{*}{Unseen} &
            False & 4.0\% & \cellcolor{blue!15}\textbf{26.0}\% & 10.0\% \\ 
        & & True & 8.0\% & 28.0\%  & \cellcolor{blue!15}\textbf{32.0}\% \\
        
    \midrule
    
    \multirow{4}{*}{\textit{Click Bell}} & 
        \multirow{2}{*}{Seen} &
            False & \cellcolor{blue!15}\textbf{36.0}\% & 24.0\% & / \\ 
        & & True & 20.0\% & \cellcolor{blue!15}\textbf{56.0}\%  & / \\
        
        \cmidrule{2-6}
        
        & \multirow{2}{*}{Unseen} &
            False & 12.0\% & \cellcolor{blue!15}\textbf{24.0}\% & 20.0\% \\ 
        & & True & 16.0\% & \cellcolor{blue!15}\textbf{48.0}\% & 14.0\% \\
    \cmidrule[\heavyrulewidth]{1-6} 
    \multirow{4}{*}{\textbf{\textit{Average}}} & 
        \multirow{2}{*}{Seen} &
            False & 29.5\% & \cellcolor{blue!15}$\textbf{43.0\%}_{\textcolor{ForestGreen}{+13.5\%}}$ & / \\ 
        & & True & 14.0\% & \cellcolor{blue!15}$\textbf{41.5\%}_{\textcolor{ForestGreen}{+27.5\%}}$  & / \\
        
        \cmidrule{2-6}
        
        & \multirow{2}{*}{Unseen} &
            False & 15.5\% & \cellcolor{blue!15}$\textbf{39.0\%}_{\textcolor{ForestGreen}{+23.5\%}}$ & ${36.5\%}_{\textcolor{ForestGreen}{+21.0\%}}$ \\ 
        & & True & 9.0\% & \cellcolor{blue!15}$\textbf{42.0\%}_{\textcolor{ForestGreen}{+33.0\%}}$ & ${29.5\%}_{\textcolor{ForestGreen}{+20.5\%}} $\\
    \cmidrule[\heavyrulewidth]{1-6} 
    \end{tabular}
    }
  \label{tab:real-world-results}%
\end{table*}

The experimental results show that real-world bimanual policies augmented with RoboTwin 2.0 achieve clear gains in robustness. In the few-shot setting—where 1,000 domain-randomized synthetic trajectories are combined with just 10 real-world demonstrations—the average success rate across all evaluation settings improves by 24.4\%, with per-configuration gains of 13.5\%, 27.5\%, 23.5\%, and 33.0\%, respectively. In the zero-shot setting trained solely on synthetic data, we still observe notable improvements of 21.0\% and 20.5\% on the two unseen-background scenarios. Notably, performance gains become larger in visually complex scenes, indicating that RoboTwin 2.0 is especially effective under challenging conditions.

These improvements stem from two factors: (1) the high visual and physical fidelity of RoboTwin 2.0, enabling direct sim-to-real transfer, and (2) the ability of domain-randomized synthetic data to prepare policies for environmental variations absent from clean real-world demonstrations. Importantly, the strong performance of the few-shot setting suggests that only minimal real-world data is needed to effectively bridge the sim-to-real gap.

\subsection{RoboTwin 2.0 Benchmark}


To evaluate the benchmarking utility and generalization challenges of RoboTwin 2.0, we assess five policy models: ACT, DP, RDT, Pi0, and DP3. \textbf{All VLAs are fine-tuned from their released pretrained weights in the single-task setting.} Evaluations are conducted on all 50 benchmark tasks using the Aloha AgileX dual-arm embodiment. For each task, \textbf{50 clean expert demonstrations} are used for training, and policies are tested with 100 rollouts under two conditions: \textit{Easy} (clean) and \textit{Hard} (domain-randomized with clutter, lighting, textures, and height variations). We provide a visualization of the benchmark setting in Appendix~\ref{benchmark_vis}. We report success rates as indicators of few-shot adaptation and robustness. Detailed setups are provided in Appendix~\ref{domain_randomization_setting} and \ref{details}, and full results are available in Appendix~\ref{full-benchmark-section} and on the \href{http://robotwin-platform.github.io/leaderboard}{Leaderboard}.

\begin{table*}[h]
  \centering
  \footnotesize
  \setlength{\tabcolsep}{6pt}
  \caption{\textbf{Subset of RoboTwin 2.0 benchmark}. Full results in Appendix~\ref{full-benchmark-section} and \href{https://robotwin-platform.github.io/leaderboard}{Leaderboard}.}
  \begin{tabular}{*{1}{>{\centering\arraybackslash}m{3.2cm}} *{10}{>{\centering\arraybackslash}m{0.48cm}}}
    \toprule
    \textbf{\makecell[c]{Simulation Task}} 
      & \multicolumn{2}{c}{\textbf{RDT}} 
      & \multicolumn{2}{c}{\textbf{Pi0}} 
      & \multicolumn{2}{c}{\textbf{ACT}} 
      & \multicolumn{2}{c}{\textbf{DP}} 
      & \multicolumn{2}{c}{\textbf{DP3}} \\
    & \textbf{Easy} & \textbf{Hard} 
    & \textbf{Easy} & \textbf{Hard} 
    & \textbf{Easy} & \textbf{Hard} 
    & \textbf{Easy} & \textbf{Hard} 
    & \textbf{Easy} & \textbf{Hard} \\
    \midrule
    \textit{Adjust Bottle} & 81\% & \cellcolor{blue!15}\textbf{75\%} & 90\% & 56\% & 97\% & 23\% & 97\% & 0\% & \cellcolor{blue!15}\textbf{99\%} & 3\% \\
    \textit{Beat Block Hammer} & \cellcolor{blue!15}\textbf{77\%} & \cellcolor{blue!15}\textbf{37\%} & 43\% & 21\% & 56\% & 3\% & 42\% & 0\% & 72\% & 8\% \\
    \textit{Blocks Ranking RGB} & 3\% & 0\% & \cellcolor{blue!15}\textbf{19\%} & \cellcolor{blue!15}\textbf{5\%} & 1\% & 0\% & 0\% & 0\% & 3\% & 0\% \\
    \textit{Blocks Ranking Size} & 0\% & 0\% & \cellcolor{blue!15}\textbf{7\%} & \cellcolor{blue!15}\textbf{1\%} & 0\% & 0\% & 1\% & 0\% & 2\% & 0\% \\
    \textit{Click Alarmclock} & 61\% & 12\% & 63\% & 11\% & 32\% & 4\% & 61\% & 5\% & \cellcolor{blue!15}\textbf{77\%} & \cellcolor{blue!15}\textbf{14\%} \\
    \textit{Click Bell} & 80\% & \cellcolor{blue!15}\textbf{9\%} & 44\% & 3\% & 58\% & 3\% & 54\% & 0\% & \cellcolor{blue!15}\textbf{90\%} & 0\% \\
    \textit{Dump Bin Bigbin} & 64\% & 32\% & 83\% & 24\% & 68\% & 1\% & 49\% & 0\% & \cellcolor{blue!15}\textbf{85\%} & \cellcolor{blue!15}\textbf{53\%} \\
    \textit{Grab Roller} & 74\% & 43\% & 96\% & \cellcolor{blue!15}\textbf{80\%} & 94\% & 25\% & \cellcolor{blue!15}\textbf{98\%} & 0\% & \cellcolor{blue!15}\textbf{98\%} & 2\% \\
    \textit{Handover Block} & 45\% & \cellcolor{blue!15}\textbf{14\%} & 45\% & 8\% & 42\% & 0\% & 10\% & 0\% & \cellcolor{blue!15}\textbf{70\%} & 0\% \\
    \textit{Handover Mic} & 90\% & \cellcolor{blue!15}\textbf{31\%} & 98\% & 13\% & 85\% & 0\% & 53\% & 0\% & \cellcolor{blue!15}\textbf{100\%} & 3\% \\
    \textit{Hanging Mug} & \cellcolor{blue!15}\textbf{23\%} & \cellcolor{blue!15}\textbf{16\%} & 11\% & 3\% & 7\% & 0\% & 8\% & 0\% & 17\% & 1\% \\
    \textit{Lift Pot} & 72\% & 9\% & 84\% & \cellcolor{blue!15}\textbf{36\%} & 88\% & 0\% & 39\% & 0\% & \cellcolor{blue!15}\textbf{97\%} & 0\% \\
     & & & &  &  & & & & & \\
     {\textbf{· · ·}} & & & &  & {\textbf{· · ·}} & & & & & \\ 
     & & & &  &  & & & & & \\
    \textit{Move Pillbottle Pad} & 8\% & 0\% & 21\% & \cellcolor{blue!15}\textbf{1\%} & 0\% & 0\% & 1\% & 0\% & \cellcolor{blue!15}\textbf{41\%} & 0\% \\    
    \midrule
    \textbf{\textit{Average (in \%)}} & 34.5 & 13.7 & 46.4 & \cellcolor{blue!15}\textbf{16.3} & 29.7 & 1.7 & 28.0 & 0.6 & \cellcolor{blue!15}\textbf{55.2} & 5.0 \\
    \bottomrule
  \end{tabular}
  \label{tab:main-benchmark}
\end{table*}

Fig.\ref{tab:main-benchmark} and Appendix~\ref{full-benchmark-section} report results on 50 tasks. Non-pretrained models (ACT, DP, DP3) perform poorly under Hard conditions, while pretrained models (RDT, Pi0) show stronger resilience, suggesting that vision–language–action pretraining provides useful priors for generalization. Still, success rates drop by 20.8\% (RDT) and 30.1\% (Pi0) from clean to randomized settings, underscoring robustness under domain shifts as a key challenge, likely due to limited diversity in pretraining data. DP3 achieves the best few-shot performance with limited samples, highlighting the role of 3D information, though its strong results partly stem from perfect point clouds and clean background segmentation in simulation. Together with Sections~\ref{good-robustness} and \ref{real-world}, these findings show RoboTwin 2.0’s value in complementing existing datasets with diverse, domain-randomized trajectories for improved generalization and robustness.

%% file: sections/related_work.tex
\section{Related Work}

\subsection{Datasets and Benchmarks for Robotic Manipulation}

Physics-based simulators underpin modern manipulation research. Existing platforms provide complementary strengths: SAPIEN~\cite{xiang2020sapien} enables dynamic interaction with 2,300+ articulated objects; ManiSkill2~\cite{maniskill2} supplies millions of demonstrations; Meta-World~\cite{metaworld}, CALVIN~\cite{mees2022calvin}, LIBERO~\cite{liu2023libero}, and RoboVerse~\cite{geng2025roboverse} target multi-task, language-conditioned, lifelong, and domain-randomized settings; RoboCasa~\cite{robocasa2024} offers large-scale human demonstrations but lacks automation and dual-arm focus.

Large-scale real-world datasets further bridge sim-to-real: AgiBot World~\cite{bu2025agibot}, RoboMIND~\cite{wu2024robomind}, Open X-Embodiment~\cite{openxembodiedment}, and Bridge~\cite{bridgedata} contribute millions of trajectories across diverse tasks, robots, and environments.

RoboTwin-1.0~\cite{mu2025robotwin} mirrored real demonstrations with simulated replicas for dual-arm benchmarking. In this work, RoboTwin 2.0 integrates LLM-driven feedback and systematic domain randomization across visual, physical, and task dimensions, producing richer corpora that improve policy robustness and generalization. A detailed comparison with prior benchmarks is provided in Appendix~\ref{benchmark-diff}.

\subsection{Robot Learning in Manipulation}

Many task-specific policy architectures \cite{wang2024rise, ke20243d, ze20243d, chi2023diffusion, fu2024mobile, chen2025g3flow,dexhanddiff, wang2024rise, adaptdiffuser,liang2024skilldiffuser,10900471,wen2025dexvla,chen2025benchmarking} achieve strong single-task performance but struggle to transfer across embodiments.
In contrast, foundation models trained on million-scale, multi-robot corpora have enabled robust zero-shot generalization: RT-1~\cite{brohan2022rt-1} unifies vision, language and actions in a single transformer for real-time kitchen tasks; RT-2~\cite{brohan2023rt-2} co-fine-tunes large vision–language models on web and robot data to unlock semantic planning and object reasoning; diffusion-based RDT-1B~\cite{liu2024rdt1b} and the $\pi_0$\cite{black2024pi_0} capture diverse bimanual dynamics from over a million episodes. Vision–language–action (VLA) frameworks like OpenVLA~\cite{openvla} and CogACT~\cite{li2024cogact}, together with adaptations like Octo~\cite{team2024octo}, LAPA~\cite{lapa}, and OpenVLA-OFT~\cite{openvla_oft} demonstrate efficient fine-tuning to novel robots and sensor modalities.

To further advance this direction, our work introduces digital-twin data collection paired with extensive domain randomization, yielding datasets that closely mirror real robot dynamics and support the training of robust and generalizable bi-manual manipulation policies.

\subsection{Domain Randomization in Imitation Learning}

Prior works have shown that randomizing visual and physical parameters, including but not limited to textures, lighting, camera pose, mass, friction and control latency combined with noise injection in expert demonstrations, enables sim-to-real transfer and robust visuomotor policies~\cite{tobin2017domain,peng2018sim,chebotar2019closing,chen2025g3flow,liang2024skilldiffuser}, and optimizing over worst-case ensembles further improves resilience to extreme domain shifts~\cite{epopt,laskey2017dart,adaptdiffuser}. However, these approaches apply randomization in isolation and lack bidirectional digital-twin feedback; our method integrates interactive simulation feedback with systematic domain randomization to generate higher-fidelity imitation data.

%% file: sections/conclusion.tex
\section{Conclusion}
\label{conclusion}
This paper presented RoboTwin 2.0, a scalable simulation framework for generating diverse, high-fidelity expert data to support robust bimanual manipulation. Our system integrates MLLM-based task generation, embodiment-adaptive behavior synthesis, and comprehensive domain randomization to address key limitations in prior synthetic data generator.

By leveraging an annotated object library and automating trajectory generation, RoboTwin 2.0 produces data with rich visual, linguistic, and physical diversity while minimizing manual engineering effort. Experiments demonstrate its effectiveness in improving policy robustness to cluttered environments, generalization to unseen tasks, and cross-embodiment manipulation.

These findings highlight the importance of scalable, automated generation of semantically rich, domain-randomized data for learning robust manipulation policies. RoboTwin 2.0 provides a foundation for unified benchmarks and scalable sim-to-real pipelines, with future work focusing on real-world deployment and multi-object task complexity.


\section{Acknowledgments}
This paper is partially supported by AgileX Robotics, D-Robotics, and the Jockey Club STEM Lab of Autonomous Intelligent Systems funded by The Hong Kong Jockey Club Charities Trust.

%% file: sections/appendix.tex
\clearpage
\appendix

\section{Contributions}
\vspace{-1em}
\begin{multicols}{2}
\noindent
\textcolor{RoboTwincolor1}{\textbf{\textit{Project Leaders}}}\\
Tianxing Chen, Yao Mu, Zhixuan Liang

\vspace{0.5em}
\noindent
\textcolor{RoboTwincolor1}{\textbf{\textit{Roadmap \& Methodology}}}\\
Yao Mu, Tianxing Chen, Ping Luo, Yusen Qin, Xiaokang Yang, Kaixuan Wang

\vspace{0.5em}
\noindent
\textcolor{RoboTwincolor1}{\textbf{\textit{Data Generator \& Benchmark}}}\\
Tianxing Chen, Zanxin Chen, Baijun Chen, Qiwei Liang, Zixuan Li, Xianliang Lin

\vspace{0.5em}
\noindent
\textcolor{RoboTwincolor1}{\textbf{\textit{CodeGen Agent}}}\\
Yibin Liu, Zanxin Chen, Yiheng Ge, Tianxing Chen, Mengkang Hu

\vspace{0.5em}
\noindent
\textcolor{RoboTwincolor1}{\textbf{\textit{RoboTwin-OD}}}\\
Baijun Chen, Qiangyu Chen, Kailun Su, Xuanbing Xie, Zanxin Chen

\vspace{0.5em}
\noindent
\textcolor{RoboTwincolor1}{\textbf{\textit{Policies Training \& Evaluation}}}\\
Tianxing Chen, Zijian Cai, Tian Nian, Huan-ang Gao, Tianling Xu


\vspace{0.5em}
\noindent
\textcolor{RoboTwincolor1}{\textbf{\textit{Real-World Deployment}}}\\
Tianxing Chen, Tian Nian, Weiliang Deng

\vspace{0.5em}
\noindent
\textcolor{RoboTwincolor1}{\textbf{\textit{Domain Randomization}}}\\
Baijun Chen, Yubin Guo, Qiwei Liang, Zhenyu Gu, Guodong Liu, Zanxin Chen, Tianxing Chen
\end{multicols} 


\vspace{-1em}

\section{Benchmarking RoboTwin 2.0 Against Existing Datasets}
\label{benchmark-diff}

We compare RoboTwin 2.0 against existing benchmarks and datasets across several key dimensions, including the number of supported tasks, the presence of domain randomization, support for automatic data generation, and compatibility with vision-language-action (VLA) model training and evaluation. The comparison is summarized in Table~\ref{tab:benchmark-comparison}.

\begin{table*}[h]
    \vspace{-5pt}
  \centering
  \footnotesize
  \caption{\textbf{Comparison of RoboTwin 2.0 with previous manipulation benchmarks and datasets.}}
  \vspace{2pt}
  \setlength{\tabcolsep}{5pt}
  \resizebox{.75\linewidth}{!}{
    \begin{tabular}{lcccc}
      \toprule
      \textbf{Benchmark \& Dataset} & \textbf{\#Tasks} & \textbf{\parbox{1.9cm}{\centering Domain\\Randomization}} & \textbf{\parbox{1.9cm}{\centering Auto Data\\Generation}} & \textbf{\parbox{1.9cm}{\centering VLA Model\\Train \& Eval}}\\
      \midrule
      Meta-world~\cite{metaworld}  & 50 & \ding{53} & \checkmark & \ding{53}\\
      Robosuite~\cite{zhu2020robosuite}  & 9 & \ding{53} & \ding{53} & \ding{53} \\
      RoboCasa~\cite{metaworld}  & 25 & \checkmark & \ding{53} & \ding{53}\\
      Maniskill2~\cite{maniskill2}  & 20 & \ding{53} & \checkmark & \ding{53} \\
      AutoBio~\cite{lan2025autobio}  & 16 & \ding{53} & \checkmark & \checkmark \\
      RoboTwin 1.0~\cite{mu2025robotwin}  & 14 & \ding{53} & \checkmark & \checkmark \\
      RoboTwin 2.0 (ours) & 50 & \checkmark & \checkmark & \checkmark\\
      \bottomrule
    \end{tabular}
  }
  \vspace{-5pt}
  \label{tab:benchmark-comparison}
\end{table*}

\vspace{-1em}

\section{Domain Randomization Setting}
\label{domain_randomization_setting}
Domain randomization in all experiments includes cluttered scenes, random lighting, table height variation (up to 3 cm), unseen language instructions and randomized background textures.

\section{Policies Training Details}
\label{details}

\textbf{RDT} in experiment~\ref{good-robustness} was pretrained for 100,000 steps with a batch size of 16 per GPU on 8 GPUs, and all single-task fine-tuning was conducted for 10,000 steps with a batch size of 16 per GPU on 4 GPUs.

\textbf{Pi0} in experiment~\ref{good-robustness} was pretrained for 100,000 steps with a batch size of 32, and all fine-tuning was performed for 30,000 steps using the same batch size.

\textbf{ACT} was trained under a unified setup with a chunk size of 50, batch size of 8, and single-GPU training for 6,000 epochs. During deployment, we applied \texttt{temporal\_agg} for temporal aggregation to improve execution stability.

\textbf{DP} was trained for 600 epochs with a batch size of 128 and a planning horizon of 8.

\textbf{DP3} was trained for 3,000 epochs with a batch size of 256, using a planning horizon of 8 and a point cloud resolution of 1,024, with precise segmentation of the background and tabletop.

\section{Support for Flexible Embodiment Combinations}
\label{embodiment-support}

Our object-centric, embodiment-agnostic data generation framework enables seamless deployment across a wide range of dual-arm robotic systems. The pipeline supports flexible embodiment configurations, allowing arbitrary combinations of heterogeneous manipulators and relative arm placements. This design ensures compatibility with diverse hardware setups and facilitates extensibility to future robotic platforms.

\begin{figure}[h] 
    \centering    \includegraphics[width=1.0\linewidth]{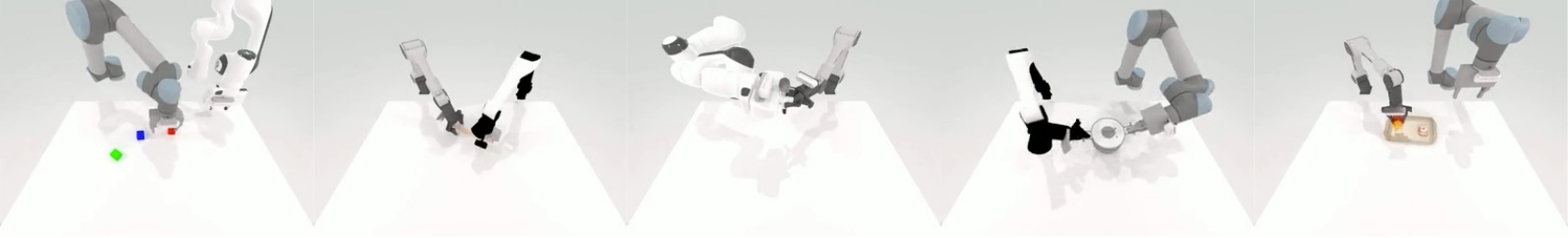}
    \vspace{-5pt}
    \caption{\textbf{Heterogeneous Dual-Arm Control via Object-Centric Manipulation.}}
    \label{fig:cross_embodiment} 
    \vspace{-5pt}
\end{figure}

To execute high-success-rate manipulation trajectories across different embodiments (see Section~\ref{grasping-adaptation}), we integrate Curobo, a high-performance, GPU-accelerated motion planner that enables efficient and reliable planning under varied kinematic constraints.

Currently, our framework supports five robotic arms—Franka, Piper, UR5, ARX-X5, and Aloha-AgileX—along with multiple gripper types, including the Panda gripper and WSG gripper. As shown in Fig.~\ref{fig:cross_embodiment}, we demonstrate successful task executions across a variety of dual-arm pairings, highlighting RoboTwin 2.0’s ability to scale to heterogeneous robot configurations and its readiness for future real-world deployment.

\section{Improvements of RoboTwin 2.0 over RoboTwin 1.0 Policy Codebase}

\begin{table}[h]
  \centering
  \begin{tabular}{lcc}
    \toprule
    \textbf{Metric} & \textbf{RoboTwin 1.0} & \textbf{RoboTwin 2.0} \\
    \midrule
    Prompt Token Length ↓       & 5901.0 & \textbf{4719.1} \\
    Code Token Length ↓         & 1236.6 & \textbf{569.4} \\
    Parallelism Control ↑       & \xmark & \checkmark \\
    AST Similarity~\cite{wen2019code} ↑ & 23.72\% & \textbf{44.78\%} \\
    CodeBLEU Similarity~\cite{ren2020codebleu} ↑ & 17.18\% & \textbf{18.53\%} \\
    CodeBERT Similarity~\cite{feng2020codebert} ↑ & 97.72\% & \textbf{98.80\%} \\
    Unixcoder Similarity~\cite{guo2022unixcoder} ↑ & 76.24\% & \textbf{82.21\%} \\
    Avg. VLM Token Cost (per observation) & -- & \textbf{6894} \\
    \bottomrule
  \end{tabular}
 \vspace{5pt}
  \caption{\textbf{Code Generation Efficiency and Quality Comparison.} Evaluation of prompt and generated code characteristics, along with code similarity metrics (AST Structural Similarity, CodeBERT, Unixcoder cosine similarity) against expert-written code, for RoboTwin 1.0 and RoboTwin 2.0 in zero-shot generation. The VLM observer cost is also reported for RoboTwin 2.0.}
\label{tab:codegen_efficiency_transposed}
\vspace{-5pt}
\end{table}

We first quantify the architectural impact of RoboTwin~2.0 in a one-shot generation without code repair and iterative refinement. Table~\ref{tab:codegen_efficiency_transposed} shows that RoboTwin~2.0 yields significantly shorter programs (569.4 vs.~1236.6 tokens), with reduced prompt length and higher structural similarity to human-written code. Crucially, it enables dual-arm parallelism via a unified API abstraction, which is absent in RoboTwin~1.0.

These improvements stem from the structured prompting and geometric API modularization designed into RoboTwin~2.0. Higher AST similarity (+21.06\%), CodeBERT similarity (+1.08\%), and Unixcoder alignment (+5.97\%) indicate that RoboTwin~2.0 not only reduces code size but also improves semantic clarity and functional alignment. 

In addition, RoboTwin~2.0 integrates a \textbf{VLM observer}, a plug-and-play module triggered only when execution fails. To quantify its overhead, we estimated VLM usage via the Kimi API (assuming each image = 1,024 tokens) over three representative tasks: the average cost was 6,295 input tokens and 599 output tokens, totaling \textbf{6,894 tokens}. While this introduces moderate overhead, the VLM enables RoboTwin~2.0 to catch and correct errors invisible to execution logging, significantly enhancing robustness and overall task success. Importantly, the observer remains optional and can be disabled when prioritizing token efficiency.

\section{Experimental Details and Metric Definitions for Code Generation}\label{code_gens}

We use the \textit{DeepSeek-V3} model for program synthesis and the \textit{moonshot-v1-32k-vision-preview} model for multimodal error localization and verification. These models were selected for their strong performance in language reasoning and visual understanding while maintaining efficiency suitable for large-scale iterative refinement.
The success rate of the $i$-th program is computed as $R_i = \frac{1}{M} \sum_{j=1}^{M} s_{i,j}$, and the final success rate for a given task under a specific system variant is then defined as $R_{\text{task}} = \frac{1}{N} \sum_{i=1}^{N} R_i$. For detailed usage instructions, refer to \url{https://robotwin-platform.github.io/doc/usage/expert-code-gen.html}.

\subsection{Metric Definitions}

We report the following metrics across all tasks:

\textbf{ASR (Average Success Rate)} is the average of $R\_{\text{task}}$ across all 10 tasks. It reflects overall task performance across all generated programs.

\textbf{Top5-ASR} is the mean success rate computed using only the top 5 highest-performing programs per task. This metric estimates system potential under a best-of-selection strategy.

\textbf{CR-Iter} indicates the average number of feedback iterations required per task before reaching a success rate above 50\% or exhausting the iteration budget.

\textbf{Token} denotes the average number of tokens of policy code generated by the language model per task. It serves as a proxy for computational cost and LLM inference budget.

These metrics jointly evaluate both the reliability and efficiency of the expert data generation pipeline under varying conditions of feedback, model capability, and refinement strategy.

\subsection{Task-Specific Performance Comparison on Code Generation}

We compare the code generation success rates of RoboTwin 2.0 and RoboTwin 1.0 across all tasks. As shown, RoboTwin 2.0 consistently matches or outperforms the baseline on the majority of tasks, demonstrating the effectiveness of our multimodal feedback and refinement pipeline.

\begin{table}[ht]
\centering
\footnotesize  
\setlength{\tabcolsep}{3.5pt}  
\begin{tabular}{lcccccc}
\toprule
Task & R1.0 Vanilla & R1.0 + FB & R1.0 + MM FB & R2.0 Vanilla & R2.0 + FB & R2.0 + MMFB \\
\midrule
beat\_block\_hammer        & 16\% & 48\% & \textbf{56\%} & 23\% & 34\% & 53\% \\
handover\_block            & 2\%  & 41\% & 45\% & 17\% & \textbf{50\%} & 27\% \\
pick\_diverse\_bottles     & \textbf{65\%} & \textbf{65\%} & 64\% & 60\% & 60\% & 62\% \\
pick\_dual\_bottles  & 99\% & 99\% & \textbf{100\%} & \textbf{100\%} & \textbf{100\%} & \textbf{100\%} \\
place\_container\_plate    & 66\% & 79\% & \textbf{91\%} & 84\% & 84\% & 82\% \\
place\_dual\_shoes         & 19\% & 22\% & \textbf{25\%} & 0\%  & 2\%  & 22\% \\
place\_empty\_cup          & 90\% & 90\% & \textbf{100\%} & 61\% & 61\% & 85\% \\
place\_shoe                & 72\% & 90\% & 90\% & \textbf{100\%} & \textbf{100\%} & \textbf{100\%} \\
stack\_blocks\_three       & 1\%  & 2\%  & 4\%  & 76\% & 76\% & \textbf{82\%} \\
stack\_blocks\_two         & 44\% & 68\% & 64\% & \textbf{100\%} & \textbf{100\%} & \textbf{100\%} \\
\bottomrule
\end{tabular}
\vspace{3pt}
\caption{
Task-Specific Performance Comparison between RoboTwin 2.0 and RoboTwin 1.0. R1.0/R2.0: RoboTwin 1.0 / 2.0. 
Bold numbers indicate the best result for each task.
}
\label{app:task_performance}
\end{table}

\subsection{Per-task Success Rates of Code Generation}

We report the success rates of all tasks in Tab.~\ref{tab:success_rates_fourcol_compact}.

\begin{table*}[h]
\caption{\textbf{Per-task success rates of our proposed R2.0 + MM FB algorithm on all RoboTwin 2.0-supported tasks.}}
\label{tab:success_rates_fourcol_compact}
\centering
\scriptsize
\setlength{\tabcolsep}{3.5pt}
\begin{tabular}{@{}ll|ll|ll|ll@{}}
\toprule
\textbf{Task} & \textbf{Rate} & \textbf{Task} & \textbf{Rate} & \textbf{Task} & \textbf{Rate} & \textbf{Task} & \textbf{Rate} \\
\midrule
Adjust Bottle & 100\% & Beat Block Hammer & 53\% & Blocks Ranking Rgb & 80\% & Blocks Ranking Size & 80\% \\
Click Alarmclock & 0\% & Click Bell & 10\% & Dump Bin Bigbin & 0\% & Grab Roller & 74\% \\
Handover Block & 27\% & Handover Mic & 0\% & Hanging Mug & 0\% & Lift Pot & 40\% \\
Move Can Pot & 30\% & Move Pillbottle Pad & 50\% & Move Playingcard Away & 90\% & Move Stapler Pad & 100\% \\
Open Laptop & 0\% & Open Microwave & 0\% & Pick Diverse Bottles & 62\% & Pick Dual Bottles & 100\% \\
Place A2B Left & 50\% & Place A2B Right & 60\% & Place Bread Basket & 0\% & Place Bread Skillet & 0\% \\
Place Can Basket & 0\% & Place Cans Plasticbox & 100\% & Place Container Plate & 82\% & Place Dual Shoes & 22\% \\
Place Empty Cup & 85\% & Place Fan & 70\% & Place Burger Fries & 100\% & Place Mouse Pad & 100\% \\
Place Object Basket & 0\% & Place Object Scale & 80\% & Place Object Stand & 90\% & Place Phone Stand & 0\% \\
Place Shoe & 100\% & Press Stapler & 0\% & Put Bottles Dustbin & 0\% & Put Object Cabinet & 0\% \\
Rotate Qrcode & 80\% & Scan Object & 0\% & Shake Bottle & 0\% & Shake Bottle Horizontally & 0\% \\
Stack Blocks Three & 82\% & Stack Blocks Two & 100\% & Stack Bowls Three & 20\% & Stack Bowls Two & 30\% \\
Stamp Seal & 20\% & Turn Switch & 0\% 
& \textit{\textcolor{RoboTwincolor1}{\textbf{Avg Success Rate}}} & \textit{\textcolor{RoboTwincolor1}{\textbf{43.34}\%}} 
&  &  \\
\bottomrule
\end{tabular}
\end{table*}

\subsection{Multimodal Observation and Error Localization}
\label{app:vlm_observe}

To further investigate the capability of the VLM observer, we manually curated a dataset of 130 execution sequences, including 101 failed trials and 29 successful trials. Each sequence consists of the natural language task instruction, a series of visual observations, and policy code. This dataset enables us to evaluate both binary error detection and fine-grained error localization.

\paragraph{Error Detection.}  
The VLM observer was first tasked with evaluating whether a robotic execution successfully completed the instructed task. The confusion matrix is as follows: TP = 16, FP = 61, TN = 40, FN = 13. The derived performance metrics are: Accuracy = 0.431, Precision = 0.208, Recall = 0.552, and F1-score = 0.302. These results suggest that while the VLM observer is reasonably sensitive in identifying execution failures (recall), its tendency to over-predict errors leads to a high false positive rate.

\paragraph{Error Localization.}  
For failed executions, we further tested whether the VLM observer could correctly localize the program step or code snippet that led to failure. Among the 40 true negative samples (correctly identified failed trials), the observer successfully pinpointed the failure step in 12 cases, yielding a localization accuracy of 30\%. This indicates that, while the observer shows promising diagnostic ability, there remains significant room for improvement in reliable error attribution.

\paragraph{Case Study.}  
We illustrate one successful case of VLM observer reasoning. Given a task instruction and five visual observations, the observer produced a step-by-step analysis:

\begin{quote}
Step 1: The left arm successfully picks up the mug from the left side. \\
Step 2: The left arm successfully places the mug at the middle position. \\
Step 3: The right arm successfully picks up the mug from the middle position. \\
Step 4: The right arm attempts to hang the mug on the rack but fails. \\
Step 5: The right arm is in the process of moving away after attempting to hang the mug. \\
The overall task was not successfully completed. The failure occurred in Step 4, where the right arm attempted to hang the mug on the rack but did not succeed. The mug is not properly positioned on the rack. The error message in the code indicates an AttributeError: 'list' object has no attribute 'p'. This error occurs in the line where the target\_point is being set to middle\_target\_pose.p. The issue is likely due to the fact that middle\_target\_pose is a list, and lists do not have an attribute 'p'. The code should be modified to correctly access the position data from the middle\_target\_pose list. \\
To fix the error, the code should be updated to access the position data correctly. For example, if middle\_target\_pose is a list containing the position and orientation, the position data should be accessed using middle\_target\_pose[0] instead of middle\_target\_pose.p. Additionally, the failure in Step 4 could be due to the incorrect position or orientation of the mug when attempting to hang it on the rack. The code should be reviewed to ensure that the mug is positioned and oriented correctly before executing the hanging motion.
\end{quote}

\paragraph{Failure Modes.}  
Despite these successes, we also observed cases where the VLM observer fails. Misclassifications primarily occur in failed executions: the observer may hallucinate correct outcomes or overlook subtle orientation errors. For instance, in a stacking task where boxes are misaligned, the observer sometimes marks the execution as successful if all boxes are placed on top of each other. Similarly, in tasks requiring fine-grained pose accuracy (e.g., placing shoes in the correct orientation), small but critical errors may be ignored. Furthermore, failures stemming from invisible factors—such as incorrect grasp axis parameters—remain challenging for purely vision-based observers to diagnose.

\subsection{LLM-Generated Code and Human-Written Code Case Study}
\label{sec:appendix_case}

\noindent\textbf{LLM-Generated Code (\texttt{gpt\_place\_shoe})}
\begin{lstlisting}[caption={LLM-Generated Code for \texttt{gpt\_place\_shoe} Task}]
class gpt_place_shoe(place_shoe):
    def play_once(self):
        # Initial observation
        self.save_camera_images(task_name="place_shoe", step_name="step1_initial_scene_state", generate_num_id="generate_num_0")
        
        # Get the shoe's position to determine which arm to use
        shoe_pose = self.shoe.get_pose()
        shoe_position = shoe_pose.p
        arm_tag = ArmTag("left" if shoe_position[0] < 0 else "right")
        
        # Grasp the shoe
        self.move(self.grasp_actor(actor=self.shoe, arm_tag=arm_tag, pre_grasp_dis=0.1, grasp_dis=0))
        self.save_camera_images(task_name="place_shoe", step_name="step2_shoe_grasped", generate_num_id="generate_num_0")
        
        # Lift the shoe up to avoid collision
        self.move(self.move_by_displacement(arm_tag=arm_tag, z=0.07, move_axis='world'))
        self.save_camera_images(task_name="place_shoe", step_name="step3_shoe_lifted", generate_num_id="generate_num_0")
        
        # Get target pose from the top functional point of the target block
        target_pose = self.target_block.get_functional_point(1, "pose")
        
        # Place the shoe on the target block with head towards left side
        self.move(self.place_actor(actor=self.shoe, arm_tag=arm_tag, target_pose=target_pose, functional_point_id=0, pre_dis=0.1, dis=0.02, is_open=True, constrain="align", pre_dis_axis='fp'))
        self.save_camera_images(task_name="place_shoe", step_name="step4_shoe_placed", generate_num_id="generate_num_0")
        
        # Lift the gripper up after placing
        self.move(self.move_by_displacement(arm_tag=arm_tag, z=0.07, move_axis='world'))
        
        # Return arm to origin
        self.move(self.back_to_origin(arm_tag=arm_tag))
        
        # Final observation
        self.move(self.save_camera_images(task_name="place_shoe", step_name="step5_final_scene_state", generate_num_id="generate_num_0"))
\end{lstlisting}

\noindent\textbf{Human-Written Code (\texttt{place\_shoe})}
\begin{lstlisting}[caption={Human-Written Code for \texttt{place\_shoe} Task}]
class place_shoe(base_task):
    def play_once(self):
        # Get the shoe's position to determine which arm to use
        shoe_pose = self.shoe.get_pose().p
        arm_tag = ArmTag("left" if shoe_pose[0] < 0 else "right")

        # Grasp the shoe with specified pre-grasp distance and gripper position
        self.move(self.grasp_actor(self.shoe, arm_tag=arm_tag, pre_grasp_dis=0.1, gripper_pos=0))

        # Lift the shoe up by 0.07 meters in z-direction
        self.move(self.move_by_displacement(arm_tag=arm_tag, z=0.07))

        # Get target_block's functional point as target pose
        target_pose = self.target_block.get_functional_point(0)

        # Place the shoe on the target_block with alignment constraint and specified pre-placement distance
        self.move(self.place_actor(self.shoe, arm_tag=arm_tag, target_pose=target_pose, functional_point_id=0, pre_dis=0.12, constrain="align"))
        
        # Open the gripper to release the shoe
        self.move(self.open_gripper(arm_tag=arm_tag))
\end{lstlisting}

The LLM generated code tends to be more verbose, explicitly logging intermediate visual states and detailing parameters (e.g., \texttt{pre\_dis\_axis='fp'}, \texttt{is\_open=True}), while human-written scripts are more minimal, omitting intermediate steps and favoring compact execution. Despite functional similarity, the structural differences illustrate that \textbf{MLLM-generated programs are not only executable but emphasize step-by-step clarity}, contributing to more robust feedback and repair.

\section{Task Instruction and Object Description Example}
\label{fig:diverse-language-demo}
\begin{tcolorbox}[
    title=Instruction Templates  (task: `Pick Dual Bottles'), 
    colback=gray!5,       
    colframe=RoboTwincolor1, 
    colbacktitle=RoboTwincolor1, 
    coltitle=white,       
]
\texttt{\scriptsize
"Use \{a\} to place \{A\} left of \{B\}.", "Set \{A\} to the left of \{B\}.", "Move \{A\} beside \{B\} using \{a\}.", "Place \{A\} on \{B\}'s left side.", "Using \{a\}, position \{A\} next to \{B\}.", "Stick \{A\} on the left of \{B\}.", "Use \{a\} and place \{A\} on \{B\}'s left.", etc
}
\end{tcolorbox}

\begin{tcolorbox}[
    title=Object Description, 
    colback=gray!5,       
    colframe=RoboTwincolor2, 
    colbacktitle=RoboTwincolor2, 
    coltitle=white,       
]
\texttt{\scriptsize
\# object id - `001\_bottle/0':\\
"red bottle",
"red soda bottle",
"plastic red bottle",
"red bottle with yellow label",
"red plastic bottle with smooth surface",
"yellow text printed on red bottle surface",
"red bottle with white label design and markings",
"red bottle with white sealing and brown top screw cap", etc\\ 
\# object id - `039\_mug/0':\\
"black mug",
"dark coffee mug",
"sleek black mug",
"black ceramic mug",
"single-handle mug",
"smooth black surface mug",
"medium-sized drinking mug",
"round mug with curved side",
"dark mug with sturdy handle",
"solid black mug with smooth finish", etc
}
\end{tcolorbox}


\section{Prompts for Generating Task Instructions and Object Descriptions}
\label{prompt-description}
\begin{lstlisting}[language=Python,caption={Prompts for Generating Task Instructions and Object Descriptions}]
# Task Instruction Template
- Goal: Generate task instruction template
- Requirements:
  - Generate 60 items. Vary in sentence length and structure
  - Use natural action verbs (grab, slide, place)
- split
  - 50 items for training
  - 10 items for evaluation

## Schema Requirements
- Goal: Use placeholders for objects in instructions
- Requirements:
  - Format: {X} for objects defined in schema
  - Include all object placeholders ({A-Z}) in every instruction
  - Omit arm references and placeholders ({a-z}) in 50% of instructions
  - Ensure natural flow when placeholders are replaced with actual values

# Object Description
- Goal: Generate natural object descriptions for robotic manipulation
- Requirements:
  - Generate 15 items. Vary in sentence length and structure
  - Use natural oral language
  - Include essential physical properties (color, shape, size, texture)
  - Use noun-focused phrases
  - For multi-part objects, use structures like `X with Y'
- split
  - 12 items for training
  - 3 items for evaluation

# Episode
An episode is a specified task, in which each task may have different objects to be manipulated,
resulting in the same task template being reused by replacing the placeholders with specific objects.
For example:
  {A} -> `medium-sized yellow bottle'
  {A} -> `green drink bottle with bold labels'

General Task -> Specific Episode:
  {A} -> bottle/0.glb
  {A} -> bottle/1.glb

The number of task instructions for an episode can be calculated by:
  Episode_num = TaskInstruction_num * ObjectDescription_num
\end{lstlisting}

\section{RoboTwin 2.0 Benchmark Setting Visualization}
\label{benchmark_vis}
\begin{figure}[h] 
    \centering    \includegraphics[width=1.0\linewidth]{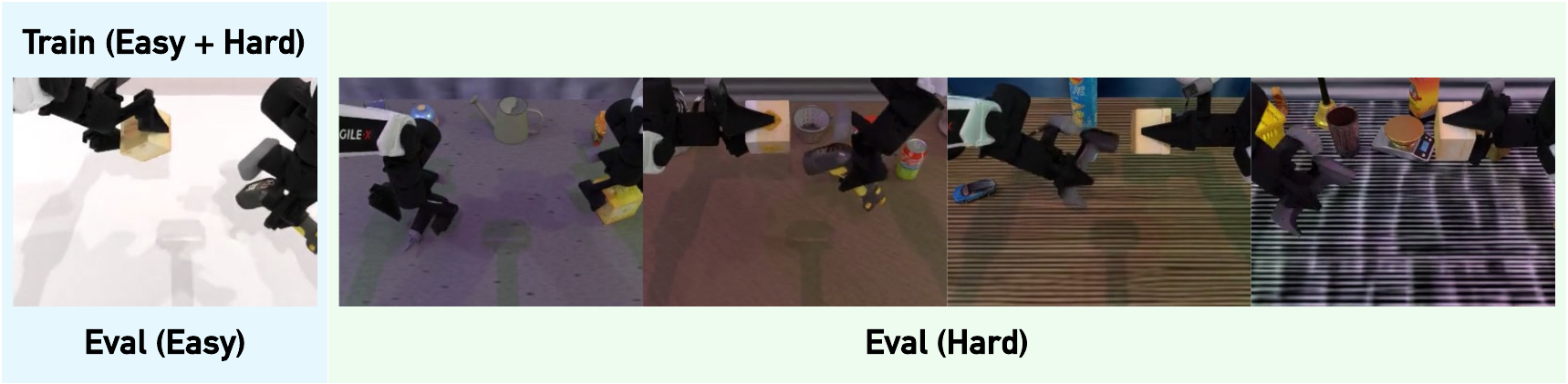}
    \vspace{-5pt}
    \caption{\textbf{Heterogeneous Dual-Arm Control via Object-Centric Manipulation.}}
    \label{fig:benchmark_vis} 
    \vspace{-5pt}
\end{figure}

We visualize the simulation settings of the RoboTwin 2.0 benchmark in Fig.~\ref{fig:benchmark_vis}. All models are trained on 50 clean (non-randomized) demonstrations per task (blue). For evaluation, the \textit{Easy} setting also uses clean environments, while the \textit{Hard} setting employs domain-randomized environments (green).

\newpage
\section{Full RoboTwin 2.0 Benchmark}
\label{full-benchmark-section}

We report the evaluation results of five policies on the RoboTwin 2.0 benchmark under the \textit{Easy} and \textit{Hard} settings. Note that these two settings differ only in evaluation conditions, while the training setup remains identical. A continuously maintained online leaderboard is available at \href{https://robotwin-platform.github.io/leaderboard}{https://robotwin-platform.github.io/leaderboard}.
\begin{table*}[h]
  \centering
  \footnotesize
  \setlength{\tabcolsep}{8pt}
  \caption{\textbf{RoboTwin 2.0 Simulation Benchmark (clean vs randomized, 50+ tasks).}}
  \begin{tabular}{*{1}{>{\centering\arraybackslash}m{3.2cm}} *{10}{>{\centering\arraybackslash}m{0.48cm}}}
    \toprule
    \textbf{\makecell[c]{Simulation Task}} 
      & \multicolumn{2}{c}{\textbf{RDT}} 
      & \multicolumn{2}{c}{\textbf{Pi0}} 
      & \multicolumn{2}{c}{\textbf{ACT}} 
      & \multicolumn{2}{c}{\textbf{DP}} 
      & \multicolumn{2}{c}{\textbf{DP3}} \\
    & \textbf{Easy} & \textbf{Hard} 
    & \textbf{Easy} & \textbf{Hard} 
    & \textbf{Easy} & \textbf{Hard} 
    & \textbf{Easy} & \textbf{Hard} 
    & \textbf{Easy} & \textbf{Hard} \\
    \midrule
    \textit{Adjust Bottle} & 81\% & \cellcolor{blue!15}\textbf{75\%} & 90\% & 56\% & 97\% & 23\% & 97\% & 0\% & \cellcolor{blue!15}\textbf{99\%} & 3\% \\
    \textit{Beat Block Hammer} & \cellcolor{blue!15}\textbf{77\%} & \cellcolor{blue!15}\textbf{37\%} & 43\% & 21\% & 56\% & 3\% & 42\% & 0\% & 72\% & 8\% \\
    \textit{Blocks Ranking RGB} & 3\% & 0\% & \cellcolor{blue!15}\textbf{19\%} & \cellcolor{blue!15}\textbf{5\%} & 1\% & 0\% & 0\% & 0\% & 3\% & 0\% \\
    \textit{Blocks Ranking Size} & 0\% & 0\% & \cellcolor{blue!15}\textbf{7\%} & \cellcolor{blue!15}\textbf{1\%} & 0\% & 0\% & 1\% & 0\% & 2\% & 0\% \\
    \textit{Click Alarmclock} & 61\% & 12\% & 63\% & 11\% & 32\% & 4\% & 61\% & 5\% & \cellcolor{blue!15}\textbf{77\%} & \cellcolor{blue!15}\textbf{14\%} \\
    \textit{Click Bell} & 80\% & \cellcolor{blue!15}\textbf{9\%} & 44\% & 3\% & 58\% & 3\% & 54\% & 0\% & \cellcolor{blue!15}\textbf{90\%} & 0\% \\
    \textit{Dump Bin Bigbin} & 64\% & 32\% & 83\% & 24\% & 68\% & 1\% & 49\% & 0\% & \cellcolor{blue!15}\textbf{85\%} & \cellcolor{blue!15}\textbf{53\%} \\
    \textit{Grab Roller} & 74\% & 43\% & 96\% & \cellcolor{blue!15}\textbf{80\%} & 94\% & 25\% & \cellcolor{blue!15}\textbf{98\%} & 0\% & \cellcolor{blue!15}\textbf{98\%} & 2\% \\
    \textit{Handover Block} & 45\% & \cellcolor{blue!15}\textbf{14\%} & 45\% & 8\% & 42\% & 0\% & 10\% & 0\% & \cellcolor{blue!15}\textbf{70\%} & 0\% \\
    \textit{Handover Mic} & 90\% & \cellcolor{blue!15}\textbf{31\%} & 98\% & 13\% & 85\% & 0\% & 53\% & 0\% & \cellcolor{blue!15}\textbf{100\%} & 3\% \\
    \textit{Hanging Mug} & \cellcolor{blue!15}\textbf{23\%} & \cellcolor{blue!15}\textbf{16\%} & 11\% & 3\% & 7\% & 0\% & 8\% & 0\% & 17\% & 1\% \\
    \textit{Lift Pot} & 72\% & 9\% & 84\% & \cellcolor{blue!15}\textbf{36\%} & 88\% & 0\% & 39\% & 0\% & \cellcolor{blue!15}\textbf{97\%} & 0\% \\
    \textit{Move Can Pot} & 25\% & 12\% & 58\% & \cellcolor{blue!15}\textbf{21\%} & 22\% & 4\% & 39\% & 0\% & \cellcolor{blue!15}\textbf{70\%} & 6\% \\
    \textit{Move Pillbottle Pad} & 8\% & 0\% & 21\% & \cellcolor{blue!15}\textbf{1\%} & 0\% & 0\% & 1\% & 0\% & \cellcolor{blue!15}\textbf{41\%} & 0\% \\
    \textit{Move Playingcard Away} & 43\% & 11\% & 53\% & \cellcolor{blue!15}\textbf{22\%} & 36\% & 0\% & 47\% & 0\% & \cellcolor{blue!15}\textbf{68\%} & 3\% \\
    \textit{Move Stapler Pad} & 2\% & 0\% & 0\% & \cellcolor{blue!15}\textbf{2\%} & 0\% & 0\% & 1\% & 0\% & \cellcolor{blue!15}\textbf{12\%} & 0\% \\
    \textit{Open Laptop} & 59\% & 32\% & \cellcolor{blue!15}\textbf{85\%} & \cellcolor{blue!15}\textbf{46\%} & 56\% & 0\% & 49\% & 0\% & 82\% & 7\% \\
    \textit{Open Microwave} & 37\% & 20\% & 80\% & \cellcolor{blue!15}\textbf{50\%} & \cellcolor{blue!15}\textbf{86\%} & 0\% & 5\% & 0\% & 61\% & 22\% \\
    \textit{Pick Diverse Bottles} & 2\% & 0\% & 27\% & \cellcolor{blue!15}\textbf{6\%} & 7\% & 0\% & 6\% & 0\% & \cellcolor{blue!15}\textbf{52\%} & 1\% \\
    \textit{Pick Dual Bottles} & 42\% & \cellcolor{blue!15}\textbf{13\%} & 57\% & 12\% & 31\% & 0\% & 24\% & 0\% & \cellcolor{blue!15}\textbf{60\%} & 1\% \\
    \textit{Place A2B Left} & 3\% & 1\% & 31\% & 1\% & 1\% & 0\% & 2\% & 0\% & \cellcolor{blue!15}\textbf{46\%} & \cellcolor{blue!15}\textbf{2\%} \\
    \textit{Place A2B Right} & 1\% & 1\% & 27\% & \cellcolor{blue!15}\textbf{6\%} & 0\% & 0\% & 13\% & 0\% & \cellcolor{blue!15}\textbf{49\%} & 0\% \\
    \textit{Place Bread Basket} & 10\% & 2\% & 17\% & \cellcolor{blue!15}\textbf{4\%} & 6\% & 0\% & 14\% & 0\% & \cellcolor{blue!15}\textbf{26\%} & 1\% \\
    \textit{Place Bread Skillet} & 5\% & \cellcolor{blue!15}\textbf{1\%} & \cellcolor{blue!15}\textbf{23\%} & \cellcolor{blue!15}\textbf{1\%} & 7\% & 0\% & 11\% & 0\% & 19\% & 0\% \\
    \textit{Place Burger Fries} & 50\% & \cellcolor{blue!15}\textbf{27\%} & \cellcolor{blue!15}\textbf{80\%} & 4\% & 49\% & 0\% & 72\% & 0\% & 72\% & 18\% \\
    \textit{Place Can Basket} & 19\% & \cellcolor{blue!15}\textbf{6\%} & 41\% & 5\% & 1\% & 0\% & 18\% & 0\% & \cellcolor{blue!15}\textbf{67\%} & 2\% \\
    \textit{Place Cans Plasticbox} & 6\% & \cellcolor{blue!15}\textbf{5\%} & 34\% & 2\% & 16\% & 0\% & 40\% & 0\% & \cellcolor{blue!15}\textbf{48\%} & 3\% \\
    \textit{Place Container Plate} & 78\% & 17\% & \cellcolor{blue!15}\textbf{88\%} & \cellcolor{blue!15}\textbf{45\%} & 72\% & 1\% & 41\% & 0\% & 86\% & 1\% \\
    \textit{Place Dual Shoes} & 4\% & \cellcolor{blue!15}\textbf{4\%} & \cellcolor{blue!15}\textbf{15\%} & 0\% & 9\% & 0\% & 8\% & 0\% & 13\% & 0\% \\
    \textit{Place Empty Cup} & 56\% & 7\% & 37\% & \cellcolor{blue!15}\textbf{11\%} & 61\% & 0\% & 37\% & 0\% & \cellcolor{blue!15}\textbf{65\%} & 1\% \\
    \textit{Place Fan} & 12\% & 2\% & 20\% & \cellcolor{blue!15}\textbf{10\%} & 1\% & 0\% & 3\% & 0\% & \cellcolor{blue!15}\textbf{36\%} & 1\% \\
    \textit{Place Mouse Pad} & 1\% & 0\% & \cellcolor{blue!15}\textbf{7\%} & \cellcolor{blue!15}\textbf{1\%} & 0\% & 0\% & 0\% & 0\% & 4\% & \cellcolor{blue!15}\textbf{1\%} \\
    \textit{Place Object Basket} & 33\% & \cellcolor{blue!15}\textbf{17\%} & 16\% & 2\% & 15\% & 0\% & 15\% & 0\% & \cellcolor{blue!15}\textbf{65\%} & 0\% \\
    \textit{Place Object Scale} & 1\% & \cellcolor{blue!15}\textbf{0\%} & \cellcolor{blue!15}\textbf{10\%} & \cellcolor{blue!15}\textbf{0\%} & 0\% & \cellcolor{blue!15}\textbf{0\%} & 1\% & \cellcolor{blue!15}\textbf{0\%} & 15\% & \cellcolor{blue!15}\textbf{0\%} \\
    \textit{Place Object Stand} & 15\% & 5\% & 36\% & \cellcolor{blue!15}\textbf{11\%} & 1\% & 0\% & 22\% & 0\% & \cellcolor{blue!15}\textbf{60\%} & 0\% \\
    \textit{Place Phone Stand} & 15\% & 6\% & 35\% & \cellcolor{blue!15}\textbf{7\%} & 2\% & 0\% & 13\% & 0\% & \cellcolor{blue!15}\textbf{44\%} & 2\% \\
    \textit{Place Shoe} & 35\% & \cellcolor{blue!15}\textbf{7\%} & 28\% & 6\% & 5\% & 0\% & 23\% & 0\% & \cellcolor{blue!15}\textbf{58\%} & 2\% \\
    \textit{Press Stapler} & 41\% & 24\% & 62\% & \cellcolor{blue!15}\textbf{29\%} & 31\% & 6\% & 6\% & 0\% & \cellcolor{blue!15}\textbf{69\%} & 3\% \\
    \textit{Put Bottles Dustbin} & 21\% & 4\% & 54\% & 13\% & 27\% & 1\% & 22\% & 0\% & \cellcolor{blue!15}\textbf{60\%} & \cellcolor{blue!15}\textbf{21\%} \\
    \textit{Put Object Cabinet} & 33\% & \cellcolor{blue!15}\textbf{18\%} & 68\% & \cellcolor{blue!15}\textbf{18\%} & 15\% & 0\% & 42\% & 0\% & \cellcolor{blue!15}\textbf{72\%} & 1\% \\
    \textit{Rotate QRcode} & 50\% & 5\% & 68\% & \cellcolor{blue!15}\textbf{15\%} & 1\% & 0\% & 13\% & 0\% & \cellcolor{blue!15}\textbf{74\%} & 1\% \\
    \textit{Scan Object} & 4\% & \cellcolor{blue!15}\textbf{1\%} & 18\% & \cellcolor{blue!15}\textbf{1\%} & 2\% & 0\% & 9\% & 0\% & \cellcolor{blue!15}\textbf{31\%} & \cellcolor{blue!15}\textbf{1\%} \\
    \textit{Shake Bottle Horizontally} & 84\% & \cellcolor{blue!15}\textbf{51\%} & 99\% & \cellcolor{blue!15}\textbf{51\%} & 63\% & 4\% & 59\% & 18\% & \cellcolor{blue!15}\textbf{100\%} & 25\% \\
    \textit{Shake Bottle} & 74\% & 45\% & 97\% & \cellcolor{blue!15}\textbf{60\%} & 74\% & 10\% & 65\% & 8\% & \cellcolor{blue!15}\textbf{98\%} & 19\% \\
    \textit{Stack Blocks Three} & 2\% & \cellcolor{blue!15}\textbf{0\%} & \cellcolor{blue!15}\textbf{17\%} & \cellcolor{blue!15}\textbf{0\%} & 0\% & \cellcolor{blue!15}\textbf{0\%} & 0\% & \cellcolor{blue!15}\textbf{0\%} & 1\% & \cellcolor{blue!15}\textbf{0\%} \\
    \textit{Stack Blocks Two} & 21\% & \cellcolor{blue!15}\textbf{2\%} & \cellcolor{blue!15}\textbf{42\%} & 1\% & 25\% & 0\% & 7\% & 0\% & 24\% & 0\% \\
    \textit{Stack Bowls Three} & 51\% & 17\% & \cellcolor{blue!15}\textbf{66\%} & \cellcolor{blue!15}\textbf{24\%} & 48\% & 0\% & 63\% & 0\% & 57\% & 5\% \\
    \textit{Stack Bowls Two} & 76\% & 30\% & \cellcolor{blue!15}\textbf{91\%} & \cellcolor{blue!15}\textbf{41\%} & 82\% & 0\% & 61\% & 0\% & 83\% & 6\% \\
    \textit{Stamp Seal} & 1\% & 0\% & 3\% & \cellcolor{blue!15}\textbf{4\%} & 2\% & 0\% & 2\% & 0\% & \cellcolor{blue!15}\textbf{18\%} & 0\% \\
    \textit{Turn Switch} & 35\% & 15\% & 27\% & \cellcolor{blue!15}\textbf{23\%} & 5\% & 2\% & 36\% & 1\% & \cellcolor{blue!15}\textbf{46\%} & 8\% \\
    \midrule
    \textbf{\textit{Average (\%)}} & 34.5 & 13.7 & 46.4 & \cellcolor{blue!15}\textbf{16.3} & 29.7 & 1.7 & 28.0 & 0.6 & \cellcolor{blue!15}\textbf{55.2} & 5.0 \\
    \bottomrule
  \end{tabular}
  \label{tab:full-main-benchmark}
\end{table*}
\FloatBarrier

\newpage
\section{Success Rates of Different Embodiments on RoboTwin 2.0 Tasks}
\label{all_planning_success_rate}
Table~\ref{tab:comparison-benchmark} reports the success rates of five robot embodiments across the 50 RoboTwin 2.0 tasks, using the same set of expert programs for data generation.

\begin{table*}[ht]
    \centering
    \scriptsize
    \setlength{\tabcolsep}{1pt}
    \caption{\textbf{Success Rates of Different Embodiments on RoboTwin 2.0 Tasks.}}
    \vspace{-2pt}
    \resizebox{\textwidth}{!}{ 
    \begin{tabular}{l *{5}{>{\centering\arraybackslash}m{0.9cm}} | *{5}{>{\centering\arraybackslash}m{0.9cm}}}
    \toprule
    & \multicolumn{5}{c}{\textbf{RoboTwin1.0}} & \multicolumn{5}{c}{\textbf{RoboTwin2.0}} \\
    \textbf{Task Name} & \textbf{Aloha} & \textbf{ARX} & \textbf{Franka} & \textbf{Piper} & \textbf{UR5} & \textbf{Aloha} & \textbf{ARX} & \textbf{Franka} & \textbf{Piper} & \textbf{UR5} \\
    \midrule
    \textit{Adjust Bottle} & 92\% & 88\% & 39\% & 0\% & 7\% & 93\% & 94\% & 34\% & 0\% & 12\% \\
    \textit{Beat Block Hammer} & 68\% & 86\% & 95\% & 0\% & 86\% & 64\% & 93\% & 98\% & 15\% & 90\% \\
    \textit{Blocks Ranking Rgb} & 92\% & 98\% & 96\% & 0\% & 82\% & 96\% & 97\% & 99\% & 13\% & 53\% \\
    \textit{Blocks Ranking Size} & 90\% & 95\% & 92\% & 0\% & 60\% & 96\% & 97\% & 89\% & 7\% & 38\% \\
    \textit{Click Alarmclock} & 89\% & 99\% & 100\% & 0\% & 95\% & 92\% & 99\% & 100\% & 0\% & 95\% \\
    \textit{Click Bell} & 100\% & 100\% & 100\% & 9\% & 100\% & 100\% & 100\% & 100\% & 91\% & 100\% \\
    \textit{Dump Bin Bigbin} & 85\% & 98\% & 90\% & 0\% & 82\% & 84\% & 100\% & 84\% & 9\% & 80\% \\
    \textit{Grab Roller} & 95\% & 69\% & 99\% & 0\% & 80\% & 95\% & 69\% & 99\% & 7\% & 81\% \\
    \textit{Handover Block} & 1\% & 3\% & 0\% & 0\% & 4\% & 83\% & 81\% & 0\% & 44\% & 0\% \\
    \textit{Handover Mic} & 62\% & 80\% & 92\% & 28\% & 0\% & 87\% & 98\% & 84\% & 65\% & 14\% \\
    \textit{Hanging Mug} & 68\% & 76\% & 5\% & 0\% & 12\% & 63\% & 73\% & 11\% & 0\% & 11\% \\
    \textit{Lift Pot} & 27\% & 50\% & 24\% & 5\% & 40\% & 27\% & 50\% & 36\% & 31\% & 40\% \\
    \textit{Move Can Pot} & 18\% & 0\% & 37\% & 2\% & 4\% & 93\% & 65\% & 92\% & 96\% & 99\% \\
    \textit{Move Pillbottle Pad} & 30\% & 52\% & 15\% & 0\% & 35\% & 67\% & 90\% & 69\% & 47\% & 86\% \\
    \textit{Move Playingcard Away} & 93\% & 100\% & 100\% & 0\% & 87\% & 99\% & 100\% & 100\% & 63\% & 66\% \\
    \textit{Move Stapler Pad} & 94\% & 92\% & 88\% & 0\% & 95\% & 92\% & 96\% & 89\% & 13\% & 75\% \\
    \textit{Open Laptop} & 76\% & 91\% & 78\% & 14\% & 55\% & 82\% & 92\% & 77\% & 23\% & 51\% \\
    \textit{Open Microwave} & 65\% & 85\% & 75\% & 5\% & 33\% & 96\% & 80\% & 59\% & 2\% & 23\% \\
    \textit{Pick Diverse Bottles} & 11\% & 1\% & 0\% & 0\% & 0\% & 51\% & 2\% & 0\% & 27\% & 4\% \\
    \textit{Pick Dual Bottles} & 8\% & 3\% & 0\% & 0\% & 0\% & 92\% & 6\% & 0\% & 81\% & 7\% \\
    \textit{Place A2B Left} & 65\% & 75\% & 70\% & 0\% & 72\% & 80\% & 88\% & 64\% & 29\% & 76\% \\
    \textit{Place A2B Right} & 70\% & 68\% & 68\% & 0\% & 69\% & 81\% & 82\% & 64\% & 31\% & 66\% \\
    \textit{Place Bread Basket} & 91\% & 91\% & 69\% & 0\% & 78\% & 89\% & 88\% & 62\% & 1\% & 67\% \\
    \textit{Place Bread Skillet} & 31\% & 28\% & 42\% & 0\% & 42\% & 34\% & 26\% & 42\% & 0\% & 37\% \\
    \textit{Place Can Basket} & 47\% & 1\% & 38\% & 0\% & 11\% & 70\% & 28\% & 61\% & 0\% & 3\% \\
    \textit{Place Cans Plasticbox} & 96\% & 93\% & 98\% & 0\% & 11\% & 100\% & 96\% & 85\% & 0\% & 82\% \\
    \textit{Place Container Plate} & 86\% & 85\% & 83\% & 0\% & 82\% & 89\% & 86\% & 86\% & 37\% & 81\% \\
    \textit{Place Dual Shoes} & 73\% & 28\% & 36\% & 0\% & 40\% & 77\% & 31\% & 41\% & 1\% & 32\% \\
    \textit{Place Empty Cup} & 92\% & 100\% & 100\% & 0\% & 100\% & 92\% & 100\% & 100\% & 4\% & 100\% \\
    \textit{Place Fan} & 93\% & 96\% & 75\% & 0\% & 85\% & 95\% & 93\% & 83\% & 0\% & 65\% \\
    \textit{Place Burger Fries} & 96\% & 95\% & 85\% & 0\% & 78\% & 97\% & 98\% & 80\% & 36\% & 74\% \\
    \textit{Place Mouse Pad} & 100\% & 80\% & 99\% & 2\% & 96\% & 99\% & 89\% & 100\% & 23\% & 73\% \\
    \textit{Place Object Basket} & 68\% & 13\% & 68\% & 0\% & 30\% & 74\% & 14\% & 61\% & 0\% & 7\% \\
    \textit{Place Object Scale} & 77\% & 93\% & 94\% & 0\% & 87\% & 78\% & 92\% & 82\% & 2\% & 76\% \\
    \textit{Place Object Stand} & 90\% & 92\% & 81\% & 0\% & 90\% & 97\% & 99\% & 81\% & 9\% & 92\% \\
    \textit{Place Phone Stand} & 66\% & 78\% & 52\% & 22\% & 44\% & 66\% & 78\% & 45\% & 53\% & 49\% \\
    \textit{Place Shoe} & 87\% & 85\% & 70\% & 0\% & 97\% & 84\% & 85\% & 74\% & 7\% & 91\% \\
    \textit{Press Stapler} & 87\% & 96\% & 99\% & 0\% & 77\% & 98\% & 96\% & 100\% & 59\% & 72\% \\
    \textit{Put Bottles Dustbin} & 0\% & 0\% & 0\% & 0\% & 0\% & 71\% & 1\% & 0\% & 56\% & 0\% \\
    \textit{Put Object Cabinet} & 13\% & 56\% & 43\% & 0\% & 0\% & 14\% & 24\% & 55\% & 0\% & 0\% \\
    \textit{Rotate Qrcode} & 78\% & 83\% & 98\% & 0\% & 81\% & 75\% & 74\% & 94\% & 0\% & 67\% \\
    \textit{Scan Object} & 8\% & 13\% & 21\% & 0\% & 8\% & 4\% & 45\% & 26\% & 0\% & 19\% \\
    \textit{Shake Bottle} & 62\% & 95\% & 82\% & 1\% & 98\% & 89\% & 94\% & 85\% & 74\% & 97\% \\
    \textit{Shake Bottle Horizontally} & 64\% & 93\% & 81\% & 1\% & 97\% & 90\% & 94\% & 85\% & 74\% & 98\% \\
    \textit{Stack Blocks Three} & 98\% & 97\% & 95\% & 0\% & 83\% & 94\% & 96\% & 80\% & 0\% & 51\% \\
    \textit{Stack Blocks Two} & 99\% & 99\% & 100\% & 0\% & 94\% & 98\% & 99\% & 96\% & 2\% & 68\% \\
    \textit{Stack Bowls Three} & 27\% & 64\% & 76\% & 0\% & 76\% & 43\% & 58\% & 82\% & 0\% & 81\% \\
    \textit{Stack Bowls Two} & 63\% & 84\% & 88\% & 0\% & 94\% & 78\% & 82\% & 88\% & 4\% & 94\% \\
    \textit{Stamp Seal} & 46\% & 91\% & 95\% & 0\% & 100\% & 56\% & 91\% & 4\% & 37\% & 100\% \\
    \textit{Turn Switch} & 27\% & 3\% & 51\% & 28\% & 10\% & 74\% & 3\% & 36\% & 81\% & 10\% \\
    \midrule
\textbf{Average} & \textbf{65.3\%} & \textbf{68.8\%} & \textbf{67.6\%} & \textbf{2.3\%} & \textbf{57.7\%} & \textbf{78.8\%} & \textbf{74.2\%} & \textbf{67.2\%} & \textbf{25.1\%} & \textbf{57.1\%}\\\textbf{Difference} & / & / & / & / & / & \textbf{\textcolor{ForestGreen}{+13.5\%}} & \textbf{\textcolor{ForestGreen}{+5.4\%}} & \textbf{\textcolor{Red}{-0.4\%}} & \textbf{\textcolor{ForestGreen}{+22.8\%}} & \textbf{\textcolor{Red}{-0.6\%}}  \\

    \bottomrule
    \end{tabular}
    }
    \label{tab:comparison-benchmark}%
\end{table*}
\FloatBarrier

\newpage